\documentclass[aos,MSNbibl,rotating,dvips]{arximspdf}
\usepackage{algorithm, algorithmic}
\usepackage{dcolumn}
\usepackage{graphicx}

\doi{10.1214/12-AOS1070} 
\volume{41}
\issue{2}
\pubyear{2013}
\firstpage{401}
\lastpage{435}

\makeatletter
\newcolumntype{d}[1]{D{.}{.}{#1}}
\newcommand{\esttwo}{\mathsf{RegLocalCLGrouping}}
\newcommand{\avg}{\mathrm{avg}}
\newcommand{\nbd}{\mathcal{N}}
\newcommand{\parent}{\operatorname{Pa}}

\newcommand{\mst}{\operatorname{MST}}
\newcommand{\dist}{\operatorname{dist}}

\newcommand{\norm}[1]{\Vert #1\Vert}

\newcommand{\atanh}{\operatorname{atanh}}

\def\nn{\nonumber}

\def\hd{\widehat{d}}

\def\hG{\widehat{G}}
\def\hP{\widehat{P}}
\def\hT{\widehat{T}}
\def\hU{\widehat{U}}

\def\hbd{\widehat{\mathbf{d}}}

\def\bfd{\mathbf{d}}

\def\bfA{\mathbf{A}}
\def\bfX{\mathbf{X}}

\def\phibf{\bolds{\phi}}

\def\thetabf{\bolds{\theta}}

\def\Ac{\mathcal{A}}
\def\Cc{\mathcal{C}}
\def\Gc{\mathcal{G}}
\def\Qc{\mathcal{Q}}
\def\Xc{\mathcal{X}}

\newcommand{\eqref}[1]{(\ref{#1})}

\def\Ebb{\mathbb{E}}
\def\Nbb{\mathbb{N}}
\def\Pbb{\mathbb{P}}

\newtheorem{theorem}{Theorem}
\newtheorem{lemma}{Lemma}
\newproclaim{definition}{Definition}
\newproclaim{remarks}{Remarks}

\newcommand{\bX}{\mathbf{X}}

\newcommand{\poly}{\mathrm{poly}}

\newcommand{\Leaf}{\operatorname{Leaf}}

\newcommand{\RG}{\mathsf{RG}}
\newcommand{\quartet}{\mathsf{Quartet}}
\newcommand{\clgroup}{\mathsf{CLGrouping}}
\newcommand{\estone}{\mathsf{LocalCLGrouping}}
\newcommand{\tree}{\operatorname{tree}}

\newcommand{\perpll}{\mathrm{Perp\mbox{-}LL}}
\newcommand{\perpbic}{\mathrm{Perp\mbox{-}BIC}}
\newcommand{\predperpll}{\mathrm{Pred\mbox{-}Perp\mbox{-}LL}}
\newcommand{\predperpbic}{\mathrm{Pred\mbox{-}Perp\mbox{-}BIC}}
\newcommand{\df}{\operatorname{df}}
\newcommand{\PMI}{\operatorname{PMI}}
\newcommand{\test}{\operatorname{test}}

\newcommand{\obs}{\operatorname{obs}}
\newcommand{\pred}{\operatorname{pred}}
\newcommand{\LDA}{\operatorname{LDA}}
\newcommand{\GM}{\operatorname{GM}}
\newcommand{\BIC}{\operatorname{BIC}}
\makeatother

\begin{document}
\begin{frontmatter}

\title{Learning loopy graphical models with latent variables: Efficient
methods and guarantees}
\runtitle{Learning loopy graphical models with latent variables}

\begin{aug}
\author{\fnms{Animashree} \snm{Anandkumar}\corref{}\thanksref{t1}\ead[label=e1]{a.anandkumar@uci.edu}}
\and
\author{\fnms{Ragupathyraj} \snm{Valluvan}\thanksref{t2}\ead[label=e2]{rvalluva@uci.edu}}
\thankstext{t1}{Supported in part by NSF Award CCF-1219234,
AFOSR Award FA9550-10-1-0310 and ARO Award
W911NF-12-1-0404.}
\thankstext{t2}{Supported by the ONR Award N00014-08-1-1015.}
\runauthor{A. Anandkumar and R. Valluvan}
\affiliation{University of California, Irvine}
\address{Electrical Engineering and\\
\quad Computer Science Department\\
University of California Irvine\\
4408 Engineering Hall\\
Irvine, California 92697\\
USA\\
\printead{e1}\\
\hphantom{E-mail:\ }\printead*{e2}} 
\end{aug}

\received{\smonth{3} \syear{2012}}
\revised{\smonth{10} \syear{2012}}

%
\begin{abstract}
The problem of structure estimation in graphical models with latent
variables is considered. We characterize conditions
for tractable graph estimation and develop efficient methods with
provable guarantees. We consider models where the underlying Markov
graph is locally tree-like, and the model is in the regime of
correlation decay. For the special case of the Ising model, the number
of samples $n$ required for structural consistency of our method scales
as $n = \Omega(\theta_{\min}^{-\delta\eta(\eta+1)-2}\log p)$, where
$p$ is the number of variables, $\theta_{\min}$ is the minimum edge
potential, $\delta$ is the depth (i.e., distance from a hidden node to
the nearest observed nodes), and $\eta$ is a parameter which depends on
the bounds on node and edge potentials in the Ising model. Necessary
conditions for structural consistency under any algorithm are derived
and our method nearly matches the lower bound on sample requirements.
Further, the proposed method is practical to implement and provides
flexibility to control the number of latent variables and the cycle
lengths in the output graph.
\end{abstract}

%


\end{frontmatter}
%
\section{Introduction}\label{sec1}

Learning latent variable models from observed samples involves mainly
two tasks: discovering relationships between the observed and hidden
variables, and estimating the strength of such relationships.
One of the simplest latent variable models is the so-called \emph{latent
class model} or \emph{n\"aive Bayes model}, where the observed variables
are conditionally independent given the state of the latent factor. An
extension of these models are \emph{latent tree models} with many hidden
variables forming a tree hierarchy. Latent tree models have been
effective in modeling data in a variety of domains, such as the
evolutionary process which gave rise to the present-day species in
bio-informatics (popularly known as \emph{phylogenetics}) \cite
{Durbinbook,semple2003phylogenetics}, for financial and topic
modeling~\cite{Choi&etal10JMLR} and for modeling contextual
information for object recognition in computer vision~\cite{choicvpr10}.
Prior works on learning latent tree models (e.g., \cite
{erdos99,mossel2007distorted,Choi&etal10JMLR}), demonstrate that
latent tree models can be learned efficiently in high dimensions. In
other words, the number of samples required for consistent learning is
much smaller than the number of variables at hand.
Moreover, inference in latent tree models is computationally tractable
by means of simple algorithms such as \emph{belief propagation}.

Despite all the above advantages, the assumption of a tree structure
may be too restrictive. For instance, in an analysis of the
relationships between topics
(encoded as latent variables) and words (corresponding to observed
variables), a latent tree model posits that the words are generated
from a single topic, while, in reality there are common words across
topics. Loopy graphical models are able to capture such relationships, while
retaining many advantages of the latent tree models.


Relaxing the tree assumption leads to nontrivial challenges: in
general, learning these models is NP-hard \cite
{Karger&Srebro01SODA,Bogdanov&etalRand}, even when there are no
latent variables, and developing methods for learning such fully
observed models is itself an area of active research (e.g., \cite
{AnandkumarTanWillskyIsing11,Jalaligreedy,Ravikumar&etal08Stat}).
In this paper, we consider structure estimation in latent graphical
models Markov on \emph{locally tree-like} graphs, meaning that local
neighborhoods in the graph do not contain cycles. Learning such graphs
has many nontrivial challenges: are there parameters regimes where
these models can be learned consistently and efficiently? If so, are
there practical learning algorithms? Are learning guarantees for loopy
models comparable to those for latent trees?
How does learning depend on various graph attributes such as node
degrees, girth of the graph and so on?
We provide answers to these questions in this paper.

\subsection{Our approach and contributions}

We consider learning latent graphical Markov models on locally
tree-like graphs in the regime of correlation decay. In this regime,
there are no long-range correlations, and the local statistics converge
to a tree limit. The implication of correlation decay is immediately
clear: we can employ the available latent tree methods to learn
``local'' subgraphs consistently, as long as they do not contain any
cycles. However, a nontrivial challenge remains: how does one merge
these estimated local subgraphs (i.e., latent trees) to obtain an
overall graph estimate? Specifically, merging involves matching latent
nodes across different latent tree estimates, and it is not clear if
this can be performed in an efficient manner.

We employ a different philosophy for building locally tree-like graphs
with latent variables. We decouple the process of introducing cycles
and latent variables in the output model. We initialize a loopy graph
consisting of only the observed variables, and then iteratively add
latent variables to local neighborhoods of the graph. We establish
correctness of our method under a set of natural conditions.

We provide precise conditions for structural consistency of $\estone$
under the probably approximately correct (PAC) model of learning
(\cite{Kearns&Vaziranibook}, page~7), for general discrete models. We simplify
these conditions for the Ising model, where each node is a binary
random variable, to obtain better intuitions. We establish that for
structural consistency, the number of samples is required to scale as
$n = \Omega(\theta_{\min}^{-\delta\eta(\eta+1)-2}\log p)$, where $p$
is the number of observed variables, $\theta_{\min}$ is the minimum
edge potential, $\delta$ is the depth (i.e., graph distance from a
hidden node to the nearest observed nodes) and $\eta$ is a parameter
which depends on the minimum and maximum node and edge potentials of
the Ising model ($\eta=1$ for homogeneous models). When there are no
hidden variables $(\delta=1)$, the sample complexity is strengthened to
$n = \Omega(\theta_{\min}^{ -2}\log p)$, which matches with the best
known sample complexity for learning fully-observed Ising models \cite
{AnandkumarTanWillskyIsing11,Jalaligreedy}.

We also establish necessary conditions for any (deterministic)
algorithm to recover the graph structure and establish that $n = \Omega
( \Delta_{\min} \rho^{-1} \log p)$ samples are necessary for structural
consistency, where $\Delta_{\min}$ is the minimum degree and $\rho$ is
the fraction of observed nodes. This is comparable to the requirement
of the proposed method under uniform node sampling (i.e., selecting the
observed nodes uniformly), given by $n = \Omega(\Delta_{\max}^2 \rho^{-2} (\log p)^3)$,
where $\Delta_{\max}$ is the maximum degree in the
graph. Thus, our method is competitive with respect to the lower bound
on learning.

Our proposed method has a number of attractive features for practical
implementation: the method is amenable to parallelization which makes
it efficient on large datasets. The method provides flexibility to
control the length of cycles and the number of latent variables
introduced in the output model. The method can incorporate penalty
scores such as the Bayesian information criterion (BIC) \cite
{schwarz1978estimating} to trade-off model complexity and fidelity.
Moreover, by controlling the cycle lengths in the output model, we can
obtain models with good inference accuracy under simple algorithms such
as loopy belief propagation (LBP). Preliminary experiments on the
newsgroup dataset suggests that the method can discover intuitive
relationships efficiently, and also compares well with the popular
latent Dirichlet allocation (LDA)~\cite{blei2003latent} in terms of
topic coherence and perplexity.

\subsection{Related work}\label{secrelated}


The classical \emph{latent class models} (LCM) consists of multivariate
distributions with a single latent variable and the observed variables
are conditionally independent under each state of the latent
variable~\cite{lazarsfeld68}. Hierarchical latent class (HLC)
models~\cite{zhang04,ZhangJMLR04,Che08} generalize these models by
allowing multiple latent variables.
However, the proposed learning algorithms are based on greedy local
search in a high-dimensional space, which is computationally expensive.
Moreover, the algorithms do not have theoretical guarantees. Similar
shortcomings also hold for expectation-maximization (EM) based
approaches~\cite{elidan05,kemp08}.
Learning latent trees has been studied extensively before, mainly in
the context of phylogenetics. See \cite
{Durbinbook,semple2003phylogenetics} for a thorough overview.
Efficient algorithms with provable performance guarantees are available
(e.g., \cite
{erdos99,daskalakis06,Choi&etal10JMLR,AnandkumarEtalspectral}). Our
proposed method in this paper is inspired by~\cite{Choi&etal10JMLR}.

Works on high-dimensional graphical model selection are more recent.
The approaches can be mainly classified into two groups: local
approaches \cite
{AnandkumarTanWillskyIsing11,Jalaligreedy,Bresler&etalRand,Sanghavi&etalAllerton10}
and those based on convex optimization \cite
{Mei06,Ravikumar&etal08Arxiv,Ravikumar&etal08Stat,Chandrasekaran10latent}.
There is a general agreement that the success of these methods is
related to the presence of correlation decay in the model \cite
{Bento&Montanari09NIPS,AnandkumarTanWillskyIsing11}. This work makes
the connection explicit: it relates the extent of correlation decay
(i.e., the convergence rate to the tree limit) with the learning
efficiency for latent models on large girth graphs. An analogous study
of the effect of correlation decay for learning fully observed models
is presented in~\cite{AnandkumarTanWillskyIsing11}.

This paper is the first work to provide provable guarantees for
learning discrete latent models on loopy graphs in high dimensions
(which can also be easily be extended to Gaussian models; see remarks
following Theorem~\ref{thmestone}).
Chandrasekharan et al.~\cite{Chandrasekaran&etal10Stat} consider learning
latent Gaussian graphical models using a convex relaxation method.
However, the method cannot be easily extended to discrete models.
Moreover, the ``incoherence'' conditions required for the success of
convex methods are hard to interpret and verify in general. In
contrast, our conditions for success are transparent and based on the
presence of correlation decay in the model. Bresler et~al. \cite
{Bresler&etalRand} considers graphical model selection with hidden
variables, but proposes learning Markov graph of marginal distribution
(upon marginalizing the hidden variables) and then replacing the
cliques in the estimated graphs with hidden variables. Sample
complexity results are not provided, and the method performs poorly in
high dimensions, since it aims to estimate dense graphs.


\section{System model}\label{sec2}

\subsection{Graphical models}
A \emph{graphical model} is a family of multivariate distributions which
are Markov in accordance to a particular undirected graph
$G=(W,E)$~\cite{Lauritzenbook}, page~32. For any distribution
belonging to the model class, a random variable $X_i$ taking value in a
set $\Xc$ is associated with each node $i\in W$ in the graph.
We consider discrete graphical models where $\Xc$ is a finite set. The
set of edges $E$ captures the set of conditional independence relations
among the random variables. We say that a set of random variables $\bX_W:=\{X_i, i \in W\}$
with probability mass function (p.m.f.) $P$ is
Markov on the graph $G$ if it factorizes according to the cliques of $G$,
%
\begin{equation}
P(\mathbf{x}) = \exp \biggl(\sum_{c\in\Cc}
\theta_c(\mathbf {x}_c)- A(\thetabf ) \biggr) \qquad\forall
\mathbf{x}\in\Xc^m,
\end{equation}
where $\Cc$ is the
set of
cliques of $G$, $m:=|W|$ is the number of variables, and $\mathbf
{x}_c$ is the
set of configurations corresponding to clique $c$. The quantity
$A(\thetabf)$ is known as the \emph{log-partition function} and serves
to normalize the probability distribution. The functions $\theta_c$ are
known as \emph{potential} functions and correspond to the \emph{canonical} parameters of the exponential family.

%


A special case is the Ising model, which is the class of pairwise
distributions over binary variables $\{-1,+1\}^m$ with probability mass
function (p.m.f.) of the form
%
\begin{equation}
\label{eqnIsing} P(\mathbf{x}) = \exp \biggl(\sum_{e\in
E}
\theta_{i,j}x_i x_j+ \sum
_{i\in V} \phi_i x_i- A(\thetabf)
\biggr)\qquad \forall \mathbf{x}\in \{-1,1\}^m.
\end{equation}
We specialize some of our results to the class of
Ising models.


We consider a multivariate distribution belonging to the class of
latent graphical models in which a subset of nodes is latent or hidden.
Let $H\subset W$ denote the hidden nodes and $V= W\setminus H$ denote
the observed nodes. Our goal is to discover the presence of hidden
variables $\bfX_H$ and learn the unknown graph structure $G(W)$, given
$n$ i.i.d. samples from observed variables $\bfX_V$. Let $p:=|V|$
denote the number of observed nodes and $m:=|W|$ denote the total
number of nodes.

\subsection{Tractable graph families: Girth-constrained graphs}
%

In general, structure estimation of graphical models is NP-hard \cite
{Karger&Srebro01SODA,Bogdanov&etalRand}. We now characterize a
tractable class of models for which we can provide guarantees on graph
estimation.

We consider the family of graphs with a bound on the \emph{girth}, which
is the length of the shortest cycle in the graph. There are many graph
constructions which lead to a bound on girth. For example, the
bipartite Ramanujan graph (\cite{Chungbook2}, page~107) and the random
Cayley graphs~\cite{gamburd2009girth} have bounds on the girth.
Recently, efficient algorithms have been proposed to generate large
girth graphs efficiently~\cite{bayati2009generating}.

Although girth-constrained graphs are locally tree-like, in general,
their global structure makes them hard instances for learning.
Specifically, girth-constrained graphs have a large tree-width: it is
known\vspace*{1pt} that a graph with average degree at least $\Delta_{\avg}$ and
girth at least $g$ has a tree width as $\Omega (\frac{1}{g+1}
(\Delta_{\avg}-1)^{\lfloor(g-1)/2\rfloor} )$ \cite
{chandran2005girth}.
Thus learning is nontrivial for graphical Markov models on
girth-constrained graphs, even when there are no latent variables due
to their large tree width~\cite{Karger&Srebro01SODA}.

%
%

\subsection{Local convergence to a tree limit}\label{seccorrdecay}

This work establishes tractable learning when the graphical model
converges locally to a tree limit.
A sufficient condition for the existence of such limits is the regime
of \emph{correlation decay},\setcounter{footnote}{2}\footnote{Technically, correlation decay can
be defined in multiple ways (\cite{MezardMontanaribook}, page~520), and
the notion we use is the uniqueness or the weak spatial mixing
condition.} which refers to the property that there are no long-range
correlations in the model \cite
{Georgiibook,MezardMontanaribook,Weitz05Algo}. This regime is also
known as the \emph{uniqueness regime} since under such an assumption,
the marginal distribution at a node is asymptotically independent of
the configuration of a growing boundary.

%

We tailor the definition of correlation decay to node neighborhoods and
provide the definition below.
Given a graph $G=(W,E)$ and a distribution $P_{\bfX_W|G}$ Markov on it,
and any subset $A\subset W$, let $P_{\bfX_A|G}$ denote the marginal
distribution of variables in $A$. For some subgraph $F\subset G$, let
$P_{\bfX_A|F}$ denote the marginal distribution on $A$ obtained by
setting the potentials of edges in $G\setminus F$ to zero. Thus,
$P_{\bfX_A|F}$ is Markov on graph $F$. Let $\nbd[i;G]:=\nbd(i;G)\cup i$
denote the closed neighborhood of node $i$ in $G$. For any two sets
$A_1, A_2\subset W$, let $\dist(A_1, A_2):=\min_{i\in A_1, j \in A_2}
\dist(i,j)$ denote the minimum graph distance.\footnote{We distinguish
between the terms \emph{graph distance} and \emph{information distances}.
The former refers to the number of edges on the shortest path
connecting the two nodes on the (unweighted) graph, while the latter
refers to the quantity in \eqref{eqnmetricnew}.} Let $B_l(i)$ denote
the set of nodes within graph distance $l$ from node $i$ and $\partial
B_l(i)$ denote the boundary nodes, that is, exactly at distance $l$
from node $i$. Let $F_l(i;G):=G(B_l(i))$ denote the induced subgraph on
$B_l(i)$.
For any distributions $P, Q$, let $\norm{P-Q}_1$ denote the $\ell_1$ norm.


\begin{definition}[(Correlation decay)]\label{defcorrdecay}A
distribution $P_{\bfX
_{W}|G}$ Markov on graph $G=(W, E)$ is said to exhibit correlation
decay with a nonincreasing rate function $\zeta(\cdot)>0$ if for all
$l \in\Nbb$,
%
\begin{equation}
\norm{P_{\bfX_{A}|G}- P_{\bfX_{A}|F_l(i;G)}}_1 \leq\zeta\bigl(\dist
\bigl(A,\partial B_l(i)\bigr)\bigr)\qquad \forall i \in W, A\subset
B_l(i).\label{eqncorrdecaydef}
\end{equation}
\end{definition}

In words, the total variation distance\footnote{Recall that the total
variation distance between two probability distributions $P, Q$ on the
same alphabet is given by $\frac{1}{2}\norm{P-Q}_1$.} between the
marginal distribution of a set~$A$ of a distribution Markov on $G$ and
the corresponding distribution Markov on subgraph $F_l(i;G)$ decays as
a function of the graph distance to the boundary. This implies that for
a class of functions $\zeta(\cdot)$, the effect of graph configuration
beyond~$l$ hops from any node $i$ has a decaying effect on the local
marginal distributions.


For the class of Ising models in \eqref{eqnIsing}, the regime of
correlation decay can be explicitly characterized, in terms of the
maximum edge potential and the maximum degree of the graph, and this is
studied in Section~\ref{secIsing}.

\section{Background on latent tree models}

We first recap the results for latent tree models which will
subsequently extended to more general latent graphical models.
It is well known that tree-structured Markov distributions on a tree
$T=(W,E)$ have a special form of factorization given by
%
\begin{equation}
P({\mathbf x}_W) = \prod_{i \in W}
P_{X_i}(x_i) \prod_{(i,j)\in T}
\frac
{P_{\bfX
_{i,j}}(x_i,x_j)}{P_{X_i}(x_i) P_{X_j}(x_j)}. \label{eqntree}
\end{equation}
Comparing with general distributions, we note that tree distributions
are directly parameterized in terms of pairwise marginal distributions
on the edges. Similarly, a Markov distribution can be described on a
rooted directed tree $\stackrel{\rightarrow}{T}$ with root $r\in W$,
where the edges of $\stackrel{\rightarrow}{T}$ are directed away from
the root. Let $\parent(i)$ denote the (unique) parent of node $i\neq r$
and $P_{X_i|X_{\parent(i)}}$ denote the corresponding conditional
distribution. The Markov distribution is given by
%
\begin{equation}
P({\mathbf x}_W) = P_{X_r}(x_r) \prod
_{i \in W, i \neq r} P_{X_i|X_{\parent(i)}}(x_i|x_{\parent(i)}).
\end{equation}
A Markov model is said to be \emph{nonsingular} \cite
{steel1994recovering,mossel2005learning} if (a) for all $e\in\stackrel
{\rightarrow}{T}$, the conditional distributions satisfy $0< |\det
(P_{X_i|X_{\parent(i)}})|<1$, and (b) for all $i \in V$, $P_{X_i}(x)>0$
for all $x \in\Xc$. A nonsingular Markov model on an undirected tree
$T$ and its directed counterpart $\stackrel{\rightarrow}{T}$ are
equivalent~\cite{steel1994recovering,mossel2005learning}.
Note that nonsingularity is equivalent to positivity (i.e., bounded
potential functions) for Markov tree models. In particular, Ising
models on trees with bounded node and edge potentials are nonsingular.
This is because under positivity, there is positive probability for any
global configuration of node states which implies that the conditional
probability at a node given any of its neighbors cannot be degenerate.

Latent tree models or phylogenetic tree models are tree-structured
graphical models in which a subset of nodes are hidden or latent. Our
goal in this paper is to leverage on the techniques developed for
learning latent tree models to analyze a more general class of latent
graphical models.



\subsection{Learning latent tree models}\label{seclearnedrees}

Learning the structure of latent tree models is an extensively studied topic.
A majority of structure learning methods (known as distance based
methods) rely on the presence of an \emph{additive tree metric}.
The additive tree metric can be obtained by considering the pairwise
marginal distributions of a tree structured joint distribution. For
instance, Mossel~\cite{mossel2007distorted} considers the following
metric for discrete distributions satisfying the nonsingular
condition
%
\begin{equation}
\label{eqnmetricgeneral}d(i,j) := -\log\bigl|\det (P_{\bfX
_{i,j}})\bigr|\qquad \forall i,j\in
W.
\end{equation}
By nonsingularity assumption,
we have that $|\det(P_{\bfX_{i,j}})|>0$ for all $i,j\in W$. The
distance metric further simplifies for some special distributions, for
example, for symmetric Ising models, it is given by the negative
logarithm of the correlation between the node pair under
consideration~\cite{semple2003phylogenetics}.

\begin{figure}

\includegraphics{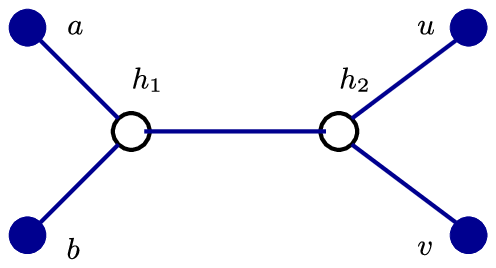}

\caption{Quartet $Q(ab|uv)$. See \protect\eqref{eqnquartet}.}\label
{figquartet}
\end{figure}

\subsubsection{Quartet-based methods}

A popular class of learning methods are based on the construction of
\emph{quartets} or \emph{splits} (e.g., \cite
{Buneman1971recovery,erdos99,mossel2007distorted}), and various
procedures to merge the inferred quartets.
A quartet is a structure over four observed nodes, as shown in Figure
\ref{figquartet}. We now recap the classical quartet test operating on
any additive tree metric. The path structure refers to the
configuration of paths between the given nodes.

\begin{definition}[(Quartet or four-point condition on trees)] Given
an additive metric on a tree $[d(i,j)]_{i,j\in V}$, the tuple of four
nodes $a,b,u,v\in V$ has the structure in Figure~\ref{figquartet} if
and only if
%
\begin{equation}
d(a,b) + d(u,v)<\min\bigl(d(a,u) + d(b,v), d(b,u) + d(a,v)\bigr),
\label{eqnquartet}
\end{equation}
and the structure in Figure \ref
{figquartet} is denoted by $Q (ab|uv)$.
\end{definition}


It is well known that the set of all quartets uniquely characterize a
latent tree. In~\cite{erdos99}, it was shown that a subset of quartets,
termed as \emph{representative quartets}, suffices to uniquely
characterize a latent tree. The set of representative quartets consists
of one quartet for each edge in the latent tree with shortest (graph)
distances between the observed nodes.


\subsubsection{Recursive grouping}
We recap the recursive grouping $\RG(\hbd^n(V),\break \Lambda,\tau)$ method
proposed in~\cite{Choi&etal10JMLR} (and its refinement in \cite
{AnandkumarEtalspectral}). The method is based on a robust quartet
test $\quartet(\hbd^n, \Lambda)$ given in Algorithm~\ref{algoquartet}.
If the confidence bound is not met, a $\perp$ result is declared. In
the first iteration of $\RG$ in Algorithm~\ref{algoRG}, the algorithm searches for node pairs
which occur on the same side of all the quartets, output by the quartet
test $\quartet(\hbd^n, \Lambda)$ and declares them as siblings and
introduces hidden variables. In later iterations of $\RG$, sibling
relationships between hidden variables are inferred through quartets
involving their children. Finally, weak edges are merged and a tree
(and more generally a forest) is output. We later use a modified
version of recursive grouping method as a routine in our algorithm for
estimating locally tree-like graphs. In the end, the neighboring nodes
(at least one of which is hidden) are merged based on the threshold
$\tau$.
See Section~\ref{secgirthalgo} for details.

\begin{algorithm}[t]
\begin{algorithmic}
\STATE\emph{Input: }Distance estimates between the observed nodes $\hbd^n(V):=\{\hd(i,j)\}_{i,j\in V}$
and confidence bound $ \Lambda$. Denote
$(\cdot)_+:=\max(\cdot,0)$.
\STATE Initialize set of quartets $\Qc(V)\leftarrow\varnothing$.
\FOR{$\{i,j,i',j'\}\in V$}
\IF{$(e^{-\hd(i,j)}-\Lambda)_+(e^{-\hd(i',j')}-\Lambda)_+
>(e^{-\hd(i,j')}+\Lambda)_+(e^{-\hd(i,j)}+\Lambda)_+$}
\STATE Declare Quartet: $\Qc(V)\leftarrow Q(ij|i'j')$.
\ENDIF
\IF{No quartet declared for $\{i,j,i',j'\}$}
\STATE$\perp_{i,j,i',j'}$ (Declare null).
\ENDIF
\ENDFOR
\end{algorithmic}
\caption{$\quartet(\hbd^n(V),\Lambda)$ test using distance estimates
$\hbd^n(V):=\{\hd(i,j)\}_{i,j\in V}$ and confidence bound $\Lambda$.}
\label{algoquartet}
\end{algorithm}

\begin{algorithm}[t]
\begin{algorithmic}
\STATE\emph{Input: }Distance estimates between the observed nodes $\hbd^n(V):=\{\hd(i,j)\}_{i,j\in V}$, confidence bound $ \Lambda$ and
threshold $\tau$. Let $\Cc(a)$ denote the children of node $a$.
\STATE Initialize $A\leftarrow V$, $\Cc(i)\leftarrow\{i\}$ for all
$i\in V$ and $\Qc(V)\leftarrow\quartet(\hbd^n(A), \Lambda)$.
\WHILE{$A\neq\varnothing$}
\IF{$\exists   i,j\in A$ s.t. for each $a\in\Cc(i)$ and $b\in\Cc
(j)$, $c,d\notin\Cc(i)\cup\Cc(j)$, $\{ac|bd, ad|bc\}\notin\Qc(V)$,
that is, $a,b$ are on same side of all such quartets in $\Qc(V)$.}
\STATE Declare $i,j$ as siblings and introduce hidden node $h$ as
parent and $\Cc(h)\leftarrow\Cc(i)\cup\Cc(j)$.
\STATE Remove $i,j$ from $A$ and add $h$ to $A$.
\ELSE
\STATE Sibling relationships cannot be further inferred. Break.
\ENDIF
\ENDWHILE
\STATE Form forest $\hT$ based on sibling and child/parent relationships.
\STATE Compute distances between any two hidden nodes as average
distance between their observed children.
\STATE Merge edges in $\hT$ of length less than $\tau$ and output
$\hT$.
\end{algorithmic}
\caption{$\RG(\hbd^n(V),\Lambda,\tau)$ test using distance estimates
$\hbd^n(V):=\{\hd(i,j)\}_{i,j\in V}$, confidence bound $\Lambda$ and
threshold $\tau$ for merging nodes.} \label{algoRG}
\end{algorithm}

\subsubsection{Chow--Liu grouping}

An alternative method, known as \emph{Chow--Liu grouping} ($\clgroup$),
was proposed in~\cite{Choi&etal10JMLR}.
Although the theoretical results for $\clgroup$ are similar to earlier
results (e.g.,~\cite{erdos99}), experiments on both synthetic and real
data sets revealed significant improvement over earlier methods in
terms of likelihood scores and number of hidden variables added.



The $\clgroup$ method always maintains a candidate tree structure and
progressively adds more hidden nodes in local neighborhoods. The
initial tree structure is the \emph{minimum spanning tree} (MST) over
the observed nodes with respect to the tree metric. The method then
considers neighborhood sets on the MST and constructs local subtrees
(using quartet based method or any other tree reconstruction
algorithm). This local reconstruction property of $\clgroup$ makes it
especially attractive for reconstructing girth-constrained graphs.

\section{Method and guarantees for structure
estimation}\label{secgirthalgo}

\subsection{Overview of algorithm}

We now describe our algorithm, which we term as $\estone$, for
structure estimation of latent graphical Markov models on graphs with
long cycles. The algorithm leverages on the Chow--Liu grouping
algorithm developed for latent tree models~\cite{Choi&etal10JMLR},
described in the previous section.\vadjust{\goodbreak} The main intuition for learning a
girth-constrained graph is based on reconstructing ``local'' parts of
the graph which are acyclic and piecing them together. However, this
approach has many challenges. First, it is not clear if the local
acyclic pieces can be learned efficiently since it requires the
presence of an additive tree metric. This is addressed by considering
models satisfying correlation decay (see Section~\ref{seccorrdecay}).
A~second and a more difficult challenge involves merging the
reconstructed local latent trees with provable guarantees due to the
introduction of unlabeled latent nodes in different pieces. We
circumvent this challenge by leveraging on the Chow--Liu grouping
algorithm~\cite{Choi&etal10JMLR} and merging the different pieces
before introducing the latent nodes.

\begin{algorithm}[t]
\begin{algorithmic}
\STATE\emph{Input: }Distance estimates between the observed nodes $\hbd^n(V):=\{\hd(i,j)\}_{i,j\in V}$, confidence bound $\Lambda$, threshold
$\tau$ and bound $r$ on distances used for local reconstruction. Let $
B_r(v;\hbd^n ):=\{u\dvtx \hd^n(u,v)\leq r\}$ and let $\mst(A;\hbd^n )$
denote the minimum spanning tree over $A\subset V$ based on edge
weights $\hbd^n(A)$. Given a graph $G$, let $\Leaf(G)$ denote the set
of nodes with unit degree. Let $\nbd[i;G]$ denote the closed
neighborhood of node $i$ in graph $G$. $\RG(\hbd^n(A), \Lambda, \tau)$
represents the recursive grouping method for building latent trees (see
Section~\ref{seclearnedrees}) over the set of nodes $A$ using distance
estimates $\hbd^n(A)$ with confidence bound $\Lambda$ and threshold
$\tau$ for merging nodes.

\FOR{$v\in V$}
\STATE$T_v\leftarrow\mst(B_r(v) ; \hbd^n)$.
\ENDFOR
\STATE Initialize $\hG, \hG_0 \leftarrow\bigcup_v T_v$.
\FOR{$v\in V\setminus\Leaf(\hG_0)$}
\STATE$A\leftarrow\nbd[v;\hG]$.
\STATE$S\leftarrow\RG( \hbd^n(A), \Lambda, \tau)$.
\STATE$\hG(A)\leftarrow S$ (Replace subgraph over $A$ with $S$ in
$\hG$)
\ENDFOR
\STATE Output $\hG$.
\end{algorithmic}
\caption{$\estone(\hbd^n(V),\Lambda, \tau,r)$ for graph estimation
using distance estimates $\hbd^n(V):=\{\hd(i,j)\}_{i,j\in V}$,
confidence bound $\Lambda$, threshold $\tau$ and distance parameter
$r$.} \label{algoscenario1}
\end{algorithm}


The algorithm is described in Algorithm~\ref{algoscenario1}.
Let $\hd^n(i,j)$ denote the estimated distance between nodes $i$ and
$j$ according to \eqref{eqnmetricgeneral} using the empirical
distribution $\hP^n_{\bfX_{i,j}}$ computed using $n$ samples, that is,
%
\begin{equation}
\hd^n(i,j) :=-\log\bigl|\det\bigl(\hP^n_{\bfX_{i,j}}\bigr)\bigr|\qquad
\forall i,j \in V.\label{eqnmetricnew}
\end{equation}
The set of distance estimates $\hbd^n(V):=\{\hd^n(i,j)\dvtx i,j\in V\}$ are input to the algorithm along with a
parameter $r$.\vadjust{\goodbreak} Recall that \mbox{$B_r(i;\hbd^n(V)):=\{j\dvtx \hd^n(i,j)\leq r\}$}.
For each observed node $i \in V$, the set of nodes $B_r(i;\hbd^n(V))$
is considered, and the minimum spanning tree is constructed. The graph
estimate $\hG$ is initialized by taking the union of all the local
minimum spanning trees.
The latent nodes are now iteratively added by considering local
neighborhoods of $\hG$ and using any latent tree algorithm for
reconstruction (e.g.,~\cite{mossel2007distorted,Choi&etal10JMLR}).
Note that the running time is polynomial (in the number of nodes) as
long as polynomial time algorithms are employed for local latent tree
reconstruction.\looseness=-1

The proposed method is efficient for practical implementation due to
the ``divide and conquer'' feature, that is, the local, latent
tree-building operations can be parallelized to obtain speedups. For
real datasets, a trade-off between model complexity and fidelity is
typically enforced by optimizing scores such as the Bayesian
information criterion (BIC)~\cite{schwarz1978estimating}. Such criteria
can be easily enforced through a greedy local search in each iteration
of our method, and this limits the number of hidden variables added by
our method. In our experiments in Section~\ref{secexperiments}, we
found that this method is quick to implement on real and synthetic datasets.

We subsequently establish the correctness of the proposed method under
a set of natural conditions. We require that the parameter $r$, which
determines the set $B_r(i;\bfd)$ for each node $i$, needs to be chosen
as a function of the depth $\delta$ (i.e., distance from a hidden node
to its closest observed nodes) and girth $g$ of the graph. In practice,
the parameter $r$ provides flexibility in tuning the length of cycles
added to the graph estimate. When $r$ is large enough, we obtain a
latent tree, while for small $r$, the graph estimate can contain many
short cycles (and potentially many components). In experiments, we
evaluate the performance of our method for different values of~$r$. The
tuning of parameters $\Lambda$ and $\tau$ has been previously discussed
in the context of learning latent trees (e.g.,~\cite{Choi&etal10JMLR}, page~1796),
and we leverage on those results here.
For more details, see Section~\ref{secexperiments}.

\begin{figure}[b]

\includegraphics{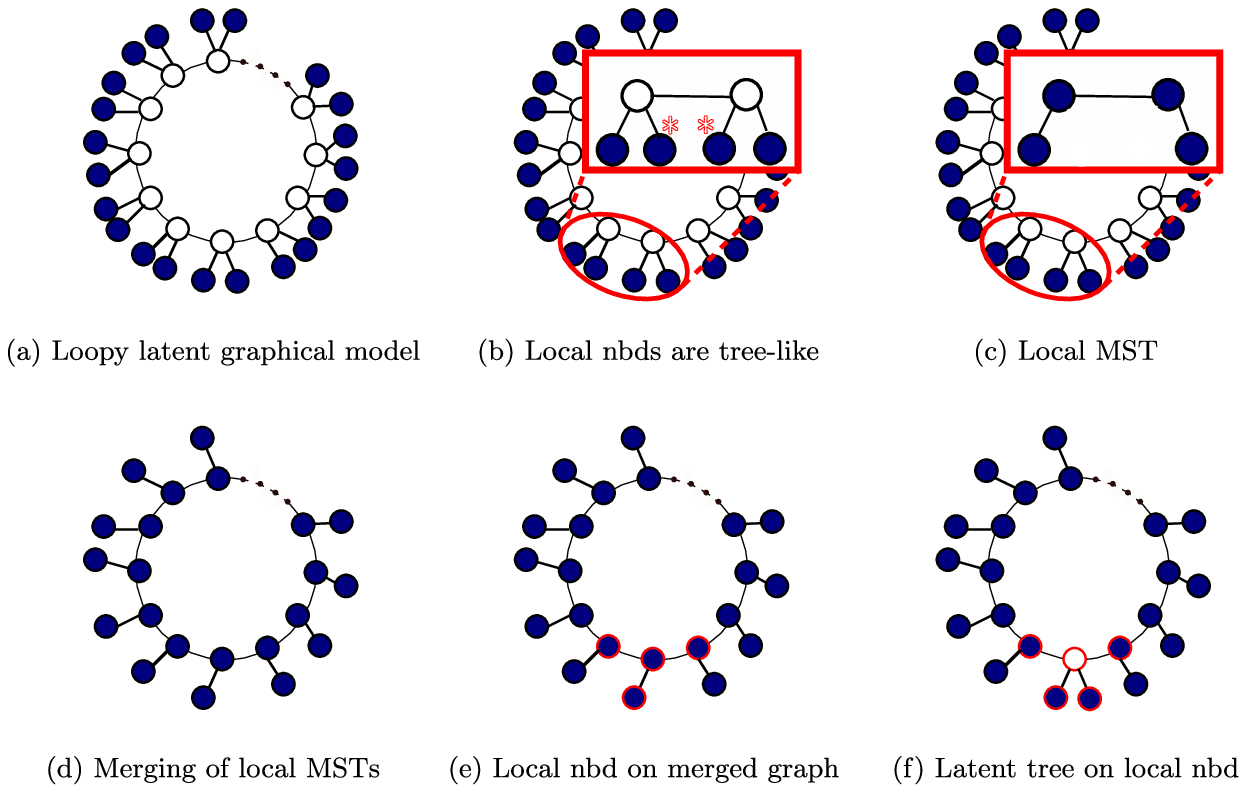}

\caption{Various steps of $\estone$ method on a simple
cycle, where observed variables are shaded. See Section \protect\ref{secexample}.}
\label{figcycle}
\end{figure}

\subsubsection{Simple example with a single cycle}\label{secexample}

To demonstrate the steps of the above proposed method, consider the
simple case of a single cycle of length $g$, where all the nodes on the
cycle are hidden, and each hidden node has two observed leaves, as
shown in Figure~\ref{figcycle}(a). When the cycle length $g $ is
sufficiently large, information distances on local neighborhoods are
approximately additive, as depicted in Figure~\ref{figcycle}(b).
Moreover, in Figure~\ref{figcycle}(b), let ``{*}'' denote the
observed node closest to each hidden node (termed as its \emph{surrogate}),
in terms of information distance. The minimum spanning
tree over the set of four nodes, which are zoomed in, corresponds to a
chain shown in Figure~\ref{figcycle}(c). Similarly, if in different
local neighborhoods of observed nodes (based on a threshold on
information distances), the surrogate relationships are similar (i.e.,
every hidden node has one of its children as its surrogate), then the
local MSTs are simple chains, and their merging gives rise to graph $G$
in Figure~\ref{figcycle}(d). Now if a local neighborhood is selected on
the merged graph $G$, as shown in Figure~\ref{figcycle}(e), then we can
discover the local latent tree structure based on information distances
as shown in Figure~\ref{figcycle}(f), since they are approximately
additive. Similarly, when different neighborhoods on~$G$ are selected,
local latent trees are discovered, and distances between nearby hidden
nodes are computed. This way we recover the latent cycle graph in
Figure~\ref{figcycle}(a) in the end.

\subsection{Results for Ising models}\label{secIsing}

We first limit ourselves to providing asymptotic guarantees for the
Ising model in \eqref{eqnIsing}, and then extend the results to
nonasymptotic guarantees in general discrete distributions.


\subsubsection{Conditions for recovery in Ising models}\label
{secconditionsIsing}

We present a set of natural conditions on the graph structure and model
parameters under which our proposed method succeeds in structure estimation.

\begin{longlist}[(A1)]

\item[(A1)] \textit{Minimum degree of latent nodes}: We require that all
latent nodes have degree at least three.


\item[(A2)]\textit{Distance bounds}: Assume bounds on the edge potentials
$\thetabf:=\{\theta_{i,j}\}$ of the Ising model
%
\begin{equation}
\label {eqnboundedge} \theta_{\min}\leq|\theta_{i,j}|\leq
\theta_{\max
} \qquad\forall (i,j)\in G.
\end{equation}
Similarly assume bounded node
potentials. We now define certain quantities which depend on the edge
potential bounds. Given a distribution belonging to the class of Ising
models $P$ with edge potentials $\thetabf=\{\theta_{i,j}\}$ and node
potentials $\phibf=\{\phi_i\}$, consider its attractive counterpart
$\bar{P}$ with edge potentials $\bar{\thetabf}:=\{|\theta_{i,j}|\}$ and
node potentials $\bar{\phibf}:=\{|\phi_i|\}$. Let $\phi'_{\max}:=
\max_{i\in V} \atanh(\bar{\Ebb}(X_i) ),$ where $\bar{\Ebb}$ is the
expectation with respect to the distribution $\bar{P}$. Let $P(\bfX_{1,2};\{\theta, \phi_1,\phi_2\})$ denote a distribution belonging to
the class of Ising models on two nodes $\{1,2\}$ with edge potential
$\theta$ and node potentials $\{\phi_1,\phi_2\}$.
Our learning guarantees depend on $d_{\min}$ and $d_{\max}$ satisfying
%
\begin{eqnarray}
d_{\min}&\geq&-\log\bigl| \det P\bigl(\bfX_{1,2};\bigl\{
\theta_{\max
}, \phi'_{\max}, \phi'_{\max}
\bigr\}\bigr)\bigr|,
\\
d_{\max} &\leq&-\log\bigl| \det P\bigl(\bfX_{1,2};\{
\theta_{\min}, 0,0\}\bigr)\bigr|,
\\
\eta&:=& \frac{d_{\max
}}{d_{\min}}.
\end{eqnarray}

\item[(A3)] \textit{Correlation decay}: We assume correlation decay in the
Ising model and require that
%
\begin{equation}
\alpha:= \Delta_{\max} \tanh \theta_{\max}<1,\qquad
\frac{\alpha^{g/2}}{\theta_{\min}^{\eta(\eta+1)+2}
}=o(1),
\end{equation}
where $\Delta_{\max}$ is the maximum node degree, $g$ is
the girth, $\theta_{\min},\theta_{\max}$ are the minimum and maximum
(absolute) edge potentials in the model and $o(1)$ is with respect to
$m$, the number of nodes in the graph.\footnote{Unless otherwise noted,
the notation $O(\cdot), o(\cdot), \Omega(\cdot), \omega(\cdot)$ are
with respect to $m$, the number of nodes in the graph.}

\item[(A4)] \textit{Girth vs. depth}: The depth $\delta$ characterizes how
close the latent nodes are to observed nodes on graph $G$: for each
hidden node $h\in H$, find a set of four observed nodes which form the
shortest \emph{quartet} with $h$ as one of the middle nodes, and
consider the largest graph distance in that quartet. The depth $\delta$
is the worst-case distance over all hidden nodes. We require the
following trade-off between the girth $g$ and the depth $\delta$,
%
\begin{equation}
\frac{g}{4} -\delta\eta (\eta+1 ) = \omega(1).\label{eqndepthgirthestone}
\end{equation}
Further, the parameter $r$ in our algorithm
is chosen as
%
\begin{equation}\label{eqnr}
r > \delta (\eta+1 )d_{\max}+\varepsilon \qquad \mbox{for some }
\varepsilon>0,\qquad \frac{g }{4}d_{\min
}- r = \omega(1).
\end{equation}
\end{longlist}

%
%

(A1) is a natural assumption on the minimum degree of the hidden
nodes for identifiability and has been imposed before for latent tree
models~\cite{Choi&etal10JMLR}. Note that the latent nodes of degree
two or lower can be marginalized to obtain an equivalent representation
of the observed statistics.

(A2) relates certain distance bounds to bounds on edge potentials.
Intuitively, $d_{\min}$ and $d_{\max}$ are bounds on information
distances given by the local tree approximation of the loopy model, and
its precise definition is given in \eqref{eqndistbound}. Note that
$e^{-d_{\max}}=\Omega(\theta_{\min})$ and $e^{-d_{\min}}=O(\theta_{\max})$.

(A3) uses bounds on the edge potentials to impose correlation decay
on the model. It is natural that the sample requirement of any graph
estimation algorithm depends on the ``weakest'' edge characterized by
the minimum edge potential $\theta_{\min}$. Further, the maximum edge
potential $\theta_{\max}$ characterizes the presence/absence of
long-range correlations in the model. Moreover, (A3) prescribes that
the extent of correlation decay be strong enough (i.e., a small $\alpha
$ and a large enough girth $g$) compared to the weakest edge in the model.

Conditions similar to (A3) have been imposed before for graphical
model selection in the regime of correlation decay when there are no
hidden variables~\cite{AnandkumarTanWillskyIsing11}. For instance,
in~\cite{AnandkumarTanWillskyIsing11}, an upper bound is imposed on
the edge potentials to limit the effect of long paths on local
conditional independence tests. A lower bound on edge potentials is
needed for edges to pass the conditional independence test.

(A4) provides the trade-off between the girth $g$ and the depth
$\delta$. Intuitively, the depth needs to be smaller than the girth to
avoid encountering cycles during the process of graph reconstruction.
Recall that the parameter $r$ in our algorithm determines the
neighborhood over which local MSTs are built in the first step. It is
chosen such that it is roughly larger than the depth $\delta$ in order
for all the hidden nodes to be discovered. The upper bound on $r$
ensures that the distortion from an additive metric is not too large.
The parameters for latent tree learning routines (such as confidence
bounds for quartet tests) are chosen appropriately depending on
$d_{\min
}$ and $d_{\max}$. See Section~\ref{secgeneral}.

\subsubsection{Guarantees for Ising models}\label{secguaranteesIsing}

We now establish that the proposed method correctly estimates the graph
structure of an Ising model in high dimensions. Recall that $\delta$ is
the depth (distance from a hidden node to its closest observed nodes),
$\theta_{\min}$ is the minimum (absolute) edge potential and $\eta
=\frac
{d_{\max}}{d_{\min}}$ is the ratio of distance bounds.

\begin{theorem}[(Structural consistency for Ising models)]\label
{thmestoneIsing}Under \textup{(A1)--(A4)}, the probability that $\estone$
is structurally consistent tends to one, when the number of samples
scales as
%
\begin{equation}
n = \Omega \bigl(\theta_{\min}^{- \delta\eta
(\eta
+1 )-2} \log p \bigr).
\label{eqnisingsample}
\end{equation}
\end{theorem}

\begin{pf}See the supplementary material~\cite{supp}.
\end{pf}

\begin{remarks*}

\begin{longlist}[(1)]
\item[(1)] For learning Ising models on locally tree-like graphs, the
sample complexity is dependent both on the minimum edge potential
$\theta_{\min}$ and on the depth $\delta$. Our method is efficient in
high dimensions since the sample requirement is only logarithmic in the
number of nodes $p$.

\item[(2)] \textit{Dependence on maximum degree}: For the correlation
decay to hold (A3), we require $\theta_{\min}\leq\theta_{\max}
=\Theta(1/\Delta_{\max})$. This implies that the sample complexity is
at least $n =\Omega(\Delta_{\max}^{\delta\eta(\eta+1)+2} \log p)$.

\item[(3)] \textit{Comparison with fully observed models}: In the special
case when all the nodes are observed $(\delta=1)$ and the graph is
locally tree-like, we strengthen the results for our method and
establish that the sample complexity for graph estimation is $n =
\Omega
(\theta_{\min}^{-2}\log p)$. This matches the best known sample
complexity for learning fully observed Ising models \cite
{AnandkumarTanWillskyIsing11,Jalaligreedy}.
The sample complexity result holds for a modified version of $\estone
$: threshold $r$ is applied to the information distances at each node
and local MSTs are formed as before. The threshold $r$ can be chosen
as
$r=d_{\max}+\varepsilon$, for some $\varepsilon>0$. The graph estimate is
obtained as the union of local MSTs and local latent tree routines are
not implemented in this case. We prove an improved sample complexity in
this special case which matches the best known bounds.

\item[(4)] \textit{Comparison with learning latent trees}: Our method is
an extension of latent tree methods for learning locally tree-like
graphs. The sample complexity of our method matches the sample
requirements for learning general latent tree models~\cite
{erdos99,mossel2007distorted,Choi&etal10JMLR}. Thus, we establish that
learning locally tree-like graphs is akin to learning latent trees in
the regime of correlation decay.
\end{longlist}
\end{remarks*}

\subsection{Extension to general discrete models}\label{secgeneral}

We now extend the results to general discrete models and provide
nonasymptotic sample requirement guarantees for success of our
proposed method.

\textit{Local tree approximation}.
We first define the notion of a local tree metric $\bfd_{\tree}(V)$
computed by limiting the model to acyclic neighborhood subgraphs
between the respective node pairs.
Given a graph $G=(W,E)$, let $\tree(i,j;G):=G(B_l(i)\cup B_l(j))$, for
$l=\lfloor g/2\rfloor-1$, denote the induced subgraph on $B_l(i)\cup
B_l(j)$, where $g$ is the girth of the graph. Recall that $B_l(i;G)$
denotes the set of nodes within graph distance $l$ from $i$ in $G$.
When $l<g/2-1$ no cycles are encountered, and thus the induced subgraph
$\tree(i,j;G)$ is acyclic. Recall that $P_{\bfX_{i,j}|G}$ denotes the
pairwise marginal distribution between $i$ and $j$ induced by the
distribution $P({\mathbf x}_W)$ Markov on graph $G$. Let $P_{\bfX
_{i,j}|\tree
(i,j)}$ denote the pairwise marginal distribution between $i$ and $j$
induced by considering only the subgraph $\tree(i,j;G)\subset G$.
Denote
%
\begin{eqnarray}
\label{eqnmetricspt} d(i,j;\tree)&:=&-\log\bigl|\det P_{\bfX
_{i,j}|\tree(i,j)}\bigr|,\nonumber\\[-8pt]\\[-8pt]
d(i,j;G)&:=&-\log\bigl| \det P_{\bfX_{i,j}|G}\bigr|.\nonumber
\end{eqnarray}
Denote $\bfd_{\tree}(V):=\{d(i,j;\tree)\dvtx i,j\in V\}$ and $\bfd(V):=\{
d(i,j;G)\dvtx i,j\in V\}$.
Note that for loopy graphs in general, $d(i,j;G)$ is different from
$d(i,j;\tree)$. The learner has access only to the empirical versions
$\hbd(V)$ of the distances $\bfd(V)$, and thus the learner cannot
estimate $\bfd_{\tree}(V)$. However, we use $\bfd_{\tree}(V)$ to
characterize the performance of our algorithm, and we list the relevant
assumptions below.


\subsubsection{Conditions on the model parameters}

\begin{longlist}[(B1)]

\item[(B1)] \textit{Minimum degree}: The minimum degree of any hidden
node in the graph is three.

\item[(B2)] \textit{Bounds on local tree metric}: Given a distribution
$P_{\bfX_W|G}$ Markov on graph~$G$, the pairwise marginal distribution
$P_{\bfX_{i,j}|\tree(i,j)}$ between any two neighbors $(i,j)\in G$ are
nonsingular and the distances
\[
d(i,j;\tree) := -\log\bigl| \det P_{\bfX
_{i,j}|\tree(i,j)}\bigr|
\]
satisfy
%
\begin{equation}
\label{eqndistbound}\quad 0<d_{\min}\leq d(i,j;\tree )\leq d_{\max
}<
\infty \qquad\forall(i,j)\in G, \qquad\eta:=\frac{d_{\max
}}{d_{\min
}}
\end{equation}
for suitable parameters $d_{\min}$ and $d_{\max}$.
%

\item[(B3)] \textit{Regime of correlation decay}: The pairwise
statistics of the distribution converge locally to a tree limit
according to Definition~\ref{defcorrdecay} with function $\zeta
(\cdot
)$ in \eqref{eqncorrdecaydef} satisfying
%
\begin{equation}
0\leq\zeta \biggl( \frac{g}{2} - \frac{r}{d_{\min}} -1 \biggr)<
\frac{\upsilon}{|\Xc|^2},\label{eqncondzeta}
\end{equation}
where $g$ is the girth, $r$ is the distance bound parameter in $\estone
$, $|\Xc|$ is the dimension of each variable, $d_{\min},d_{\max}$ are
the distance bounds in \eqref{eqndistbound} and
%
\begin{eqnarray}
\label{eqnupsilon} &&\upsilon:= \min \biggl(d_{\min
},0.5e^{-r}
\bigl(e^{d_{\min}}-1\bigr), e^{-0.5d_{\max}({r}/{d_{\min
}}+2)},
\nonumber
\\[-8pt]
\\[-8pt]
\nonumber
&&\hspace*{110pt}\frac{g }{4}d_{\min}-
r, r-d_{\max}\delta(\eta+1) \biggr).
\end{eqnarray}

\item[(B4)] \textit{Confidence bound for quartet test}: The confidence
bound in\break $\quartet(\hbd, \Lambda)$ routine in Algorithm \ref
{algoquartet} is chosen as
%
\begin{equation}
\label{eqnLambda} \Lambda= \exp \biggl[-\frac{d_{\max}}{2}\biggl(
\frac{r}{d_{\min
}}+2\biggr) \biggr].
\end{equation}

\item[(B5)]\textit{Threshold for merging nodes}: The threshold $\tau$ in
$\RG(\hbd, \Lambda, \tau)$ routine in Algorithm~\ref{algoRG} is
chosen as
%
\begin{equation}
\label{eqntau} \tau= \frac{d_{\min}}{2} - |\Xc|^2 \zeta \biggl(
\frac
{g}{2}-1\biggr)>0,
\end{equation}
where $|\Xc|$ is the dimension of the variable at each node, and
$\zeta
(\cdot)$ is the correlation decay function according to \eqref
{eqncorrdecaydef}.

\end{longlist}

(B1) is a natural assumption on the minimum degree of the hidden
nodes for identifiability, which is also needed for latent trees.
Assumption (B2) states that every edge has bounded distances under
local tree approximations. Recall that in the special case of Ising
models, this can be expressed via bounds on edge potentials. Assumption
(B3) on correlation decay imposes a constraint on the rate function
$\zeta(\cdot)$, in terms of the girth of the graph $g$, the distance
threshold $r$ used by the proposed method, the distance bounds $d_{\min
}$ and $d_{\max}$ and depth $\delta$.
(B3) implies that we require that the depth $\delta$ satisfies
%
\begin{equation}
\frac{g}{4} d_{\min} >\delta (\eta+1 )d_{\max} .\label
{eqndepthgirthestone2}
\end{equation}
Similarly, (B3) imposes constraints on
the parameter $r$ used by the proposed algorithm for building local
minimum spanning\vadjust{\goodbreak} trees in the first step. (B3) implies that~$r$ needs
to be chosen as
%
\begin{equation}
\delta (\eta+1 )d_{\max}< r <\frac{g
}{4}d_{\min}- r.
\label{eqnr2}
\end{equation}
Intuitively, the above constraint implies that $r$ is relatively small
compared to the girth of the graph and large enough for every hidden
node to be discovered. This enables the proposed algorithm to correct
reconstruct latent trees locally.

The confidence bound constraint in (B4) is based on the concentration
bounds for the empirical distances. The threshold for merging nodes in
(B5) ensures that spurious hidden nodes are not added. These
conditions are inherited from latent tree algorithms.

\subsection{Guarantees for the proposed method}

We now establish that the $\estone$ algorithm is structurally
consistent under the above conditions.

\begin{theorem}[(Structural consistency of $\estone$)]\label{thmestone}Under
assumptions \textup{(B1)--(B5)}, the $\estone$ algorithm is structurally
consistent with probability at least $1-\kappa$, for any $\kappa>0$,
when the sample size $n$ satisfies
%
\begin{equation}\quad
n > \frac{2 |\Xc
|^2}{(\upsilon-
|\Xc|^2 \zeta( {g}/{2} - {r}/{d_{\min}} -1))^2} \biggl(4\log p+|\Xc|\log2 - \log\frac{\kappa}{7}
\biggr),
\end{equation}
where $\upsilon$ is
given by \eqref{eqnupsilon}.
\end{theorem}

\begin{remarks*}

\begin{longlist}
\item[(1)] We provide PAC guarantees for reconstructing latent
graphical models on girth-constrained graphs. The conditions for
success imposed on the girth of the graph are relatively mild. We
require that the girth be roughly larger than the depth and that the
correlation decay function $\zeta(\cdot)$ be sufficiently strong
(B3). Thus, learning girth-constrained graphs is akin to learning
latent tree models (in terms of sample and computational complexities)
under a wide range of conditions.

\item[(2)] One notable additional condition required for learning
girth-con\-strained graphs in contrast to latent trees is the requirement
of correlation decay (B3). However, we note that this is only a
sufficient condition, and not necessary for learnability. For instance,
the result in~\cite{Dembo&Montanari08Arxiv} establishes that the
pairwise statistics converge locally to a tree limit for all attractive
Ising models with strictly positive node potentials, but without any
additional constraints on the parameters. Our results and analysis hold
in such scenarios since we only require local convergence to a tree metric.
\item[(3)] The results above are applicable for discrete models but can
be extended to Gaussian models using the notion of \emph{walk-summability}
in place of correlation decay according to \eqref
{eqncorrdecaydef}\vadjust{\goodbreak} (see~\cite{AnandkumarTanWillskyGaussian11}) and the
negative logarithm of the correlation coefficient as the distance
metric; see~\cite{Choi&etal10JMLR}. The results can also be extended
to more general linear models such as multivariate Gaussian models,
Gaussian mixtures and so on, along the lines of~\cite{AnandkumarEtalspectral}.
\end{longlist}
\end{remarks*}

\begin{pf*}{Proof of Theorem~\ref{thmestone}}
The detailed proof is given in the supplementary material~\cite{supp}. It
consists of the following main steps:
\begin{longlist}
\item[(1)] We first prove correctness of $\estone$ under the tree limit
[i.e., distances $\bfd_{\tree}(V):=\{d(i,j;\tree)\}_{i,j\in V}$] and
then show sample-based consistency. The latter is based on
concentration bounds, along the lines of analysis for latent tree
models~\cite{erdos99,mossel2007distorted}, with an additional
distortion introduced due to the presence of a loopy graph.

\item[(2)] We now briefly describe the proof establishing the
correctness of $\estone$ algorithm under $\bfd_{\tree}$ in
girth-constrained graphs. Intuitively, the distances $d(i,j;\tree)$
correspond to a tree metric when the graph distance $\dist(i,j)<
g/2-1$, where $g$ is the girth. Since $\estone$ infers latent trees
only locally, it avoids running into cycles and thus correctly
reconstructs the local latent trees. The initialization step in
$\estone
$ corresponds to the correct merge of this local latent trees under the
assumptions on parameter $r$ in~\eqref{eqnr2}, and the correctness of
$\estone$ is established.\quad\qed
\end{longlist}
\noqed\end{pf*}

\subsubsection{Guarantees under uniform sampling}
We have so far given guarantees for graph reconstruction, given an
arbitrary set of observed nodes in the graph. We now specialize the
results to the case when there is a uniform sampling of nodes and
provide learning guarantees. This analysis provides intuitions on the
relationship between the fraction of sampled nodes and the resulting
learning performance.

Consider an ensemble of graphs on $m$ nodes with girth at least $g$ and
minimum degree $\Delta_{\min}\geq3$ and maximum degree $\Delta_{\max
}$. Let $\rho:=\frac{p}{m}$ denote the uniform sampling probability for
selecting observed nodes. We have the following result on the depth
$\delta$. Define a constant $\varepsilon_0>0$ as
%
\begin{equation}
\varepsilon_0 = -\frac
{\log(4m \Delta_{\max}(1-\rho)^{(\Delta_{\min}-1)^{g/2}}) }{\log m}.
\end{equation}

\begin{lemma}[(Depth under uniform sampling)]\label{lemmagirthsampling}
Given uniform sampling probability of $ \rho$, for any $\varepsilon\leq
\max(0,\varepsilon_0)$,
%
\begin{equation}\quad
\delta<\frac{1}{\log(\Delta_{\min
}-1)} \biggl(\log \biggl[\frac{ \log(4 m^{1+\varepsilon}\Delta_{\max})}{|\log
(1-\rho
)|} \biggr] \biggr)\qquad
\mbox{w.p.} \geq1-m^{-\varepsilon}.\label {eqngirthsampling}
\end{equation}
\end{lemma}

\begin{pf}The proof is by straightforward arguments on binomial random
variables and the union bound. See the supplementary material~\cite{supp}.
\end{pf}

\begin{remarks*}

\begin{longlist}
\item[(1)] Assuming that the girth satisfies $g > 2\delta(1+d_{\max
}/d_{\min})$ w.h.p., when the sampling probability and the degrees are
both constant, then
%
\begin{eqnarray}
\rho=\Theta(1), \qquad\Delta_{\min},\Delta_{\max
}=O(1) \Rightarrow
\delta=O(\log\log m)\Rightarrow n =\Omega\bigl(\poly (\log m)\bigr),\nonumber\\
\eqntext{\mbox{w.h.p.}
\nn,}
\end{eqnarray}
where $\poly(\log m)$ refers to a
polylogarithmic dependence in $m$. On the other hand, with vanishing
sampling probability, for $\beta\in[0,1)$, we have
%
\begin{eqnarray}
\nn\rho=\Theta\bigl(m^{\beta-1}\bigr),\qquad \Delta_{\min
},
\Delta_{\max}=O(1) \Rightarrow\delta=O(\log m)\Rightarrow n = \Omega\bigl(
\poly(m)\bigr),\nonumber\\
\eqntext{ \mbox {w.h.p.}}
\end{eqnarray}

\item[(2)] Recall that for Ising models, the best-case sample
complexity of $\estone$ for structural consistency [when $\eta=1$ and
$\theta_{\min}=\theta_{\max}=\Theta(1/\Delta_{\max})$] scales as
\[
n = \Omega\bigl(\Delta_{\max}^{2(\delta+1)} \log p\bigr).
\]
Thus, under uniform sampling, the sample complexity required for
consistency scales as
\[
n= \Omega \biggl(\Delta_{\max}^2 \biggl(\frac{\log p}{|\log(1-\rho
)|}
\biggr)^{4{\log\Delta_{\max}}/{\log(\Delta_{\min}-1)}} \log p \biggr).
\]
For the special case when the graph is regular $(\Delta_{\min}=\Delta_{\max})$, this reduces to
%
\begin{equation}
n=\Omega \bigl(\Delta_{\max}^2 \rho^{-2} (\log
p)^3 \bigr). \label{eqnisingunif}
\end{equation}




\end{longlist}
\end{remarks*}

\section{Necessary conditions for graph estimation}

We have so far provided sufficient conditions for recovering latent
graphical Markov models on girth-constrained graphs. We now provide
necessary conditions on the number of samples required by any algorithm
to reconstruct the graph.
Let $\widehat{G}_n \dvtx (\Xc^{|V|})^n \to\Gc_m$ denote any deterministic
graph estimator using $n$ i.i.d. samples from node set $V$, and $\Gc_m$
is the set of all possible graphs on $m$ nodes.

We first define the notion of the graph edit distance based on inexact
graph matching~\cite{Bunke83}.
Let $G,\hG$ be two graphs with common labeled node set $V$ and
unlabeled node sets $U$ and $\hU$. Without loss of generality, let $|U|
\geq|\hU|$ and add $|U|-|\hU|$ number of isolated nodes to $\hG$. Let
$ \bfA_G,\bfA_{\hG}$ be the resulting adjacency matrices. Then the edit
distance between $G,\hG$ is defined as
\[
\dist(\hG,G;V):=\min_{\pi} \bigl\Vert \bfA_{\hG} - \pi(
\bfA_G)\bigr\Vert_1,
\]
where $\pi$ is any permutation on the unlabeled nodes while keeping the
labeled node set $V$ fixed.\vadjust{\goodbreak}

In other words, the edit distance is the minimum number of entries that
are different in $\bfA_{\hG}$ and in any permutation of $\bfA_G$ over
the unlabeled nodes. In our context, the labeled nodes correspond to
the observed nodes $V$ while the unlabeled nodes correspond to latent
nodes $H$. We now provide necessary conditions for graph reconstruction
up to certain edit distance.

\begin{theorem}[(Necessary condition)]\label{thmnecessary}For any deterministic
estimator $\hG_m \dvtx\break (\Xc^{m^\beta})^n\mapsto\Gc_m$ based on $n$ i.i.d.
samples from $m^\beta$ observed nodes $\beta\in[0,1]$ of a latent
graphical Markov model on graph $G_m $ on $m$ nodes with girth $g$,
minimum degree $\Delta_{\min}$ and maximum degree $\Delta_{\max}$, for
all $\varepsilon>0$, we have
%
\begin{equation}
\quad\Pbb\bigl[\dist(\hG_m,G_m;V)>\varepsilon m\bigr] \geq1-
\frac{|\Xc|^{n m^\beta
}m^{(2\varepsilon+1)m} 3^{\varepsilon m}}{ m^{0.5 \Delta_{\min} m}
(m-g\Delta_{\max}^g)^{0.5 \Delta_{\min} m}},
\end{equation}
under any sampling process used
to choose the observed nodes.
\end{theorem}

\begin{pf}The proof is based on counting arguments. See the supplementary
material~\cite{supp} for details.
\end{pf}

\begin{remarks*}

\begin{longlist}[(1)]
\item[(1)] The above result states that roughly
%
\begin{equation}
n = \Omega \bigl(\Delta_{\min} m^{1-\beta} \log m\bigr)= \Omega
\biggl(\frac{\Delta_{\min}}{
\rho} \log p \biggr)\label{eqnnecessary}
\end{equation}
samples are required for
structural consistency. Thus, when $\beta=1$ (constant fraction of
observed nodes), logarithmic number of samples are necessary while when
$\beta<1$ (vanishing fraction of observed nodes), polynomial number of
samples are necessary for reconstruction. From \eqref{eqnisingunif},
recall that for Ising models, under uniform sampling of observed nodes,
the best-case sample complexity of $\estone$ [for homogeneous models on
regular graphs with degree $\Delta$ and $\theta_{\min}=\theta_{\max}=
\Theta(1/\Delta)$] scales as
\[
n = \Omega\bigl(\Delta^2 \rho^{-2} (\log p)^3
\bigr),
\]
and thus nearly matches the lower bound on sample complexity in \eqref
{eqnnecessary}.

\end{longlist}
\end{remarks*}

\section{Experiments}\label{secexperiments}

In this section we present experimental results on real and synthetic
data. We evaluate performance in terms of perplexity, predictive
perplexity and topic coherence, used frequently in topic modeling. In
addition, we also study trade-off between model complexity and data
fitting through the Bayesian information criterion (BIC) \cite
{schwarz1978estimating}. Experiments are conducted using the
20-newsgroup data set, monthly stock returns from the S$\&$P 100
companies and synthetic data.
The datasets, software code and results are available at
\url{http://newport.eecs.uci.edu/anandkumar}.


\subsection{Experimental setup}

\textit{Synthetic data}. We generate samples from an Ising model
Markov on a cycle (see\vadjust{\goodbreak} Figure~\ref{figcycle}) with a fixed depth
$\delta= 1$, a~fixed latent node degree $\Delta= 4$ and different
girths $g = 10, 20, 30,\ldots,100$. The node potentials are kept at zero,
while the edge potentials are chosen randomly in the range $[0.05,
0.2]$. This ensures that the model remains in the regime of correlation
decay since the critical potential $\theta^{*} = \atanh( \Delta^{-1})
= 0.2554>0.2$.

\textit{Newsgroup data.} We employ latent graphical models for topic
modeling, that is, modeling the relationships between various words
co-occurring in documents. Each hidden variable in the model can be
thought of as representing a topic, and topics and words in a document
are drawn jointly from the graphical model. For a latent tree graphical
model, topics and words are constrained to form a tree, while loopy
models relax this assumption. We consider $n =16\mbox{,}242$ binary samples of
$p=100$ keywords selected from the 20 newsgroup data. Each binary
sample indicates the appearance of the given words in each posting.
These samples are divided in to two equal groups, training and test
sets for learning and testing purposes.

\textit{S$\&$P data}. We also employ latent graphical models for
financial modeling and in particular, for estimating the dependencies
between the stock trends of different companies. The data set consists
of monthly stock returns of $p=84$ companies\footnote{The 16 companies
added after 1990 are dropped from the list of 100 companies listed in
S$\&$P 100 stock index for this analysis.} listed in S$\&$P 100 index
from 1990 to 2007. Experiments with this dataset allows us to
demonstrate the performance of our algorithm on data using a Gaussian
graphical model. The Gaussian model is a simplifying assumption but
reveals interesting relationships between the companies. We note that
more sophisticated kernel models can indeed be used in place of the
Gaussian approximation, for example,~\cite{songkernel}.

This allows us to trade-off model complexity and data fitting. In
addition, we obtain better generalization by avoiding overfitting. Note
that our proposed method only deals with structure estimation and we
use expectation maximization (EM) for parameter estimation.
For the newsgroup data we compare the proposed method with the LDA
model.\footnote{Typically, LDA models the counts of different words in
documents. Here, since we have binary data, we consider a binary LDA
model where the observed variables are binary.}

\textit{Implementation}. The above method is implemented in MATLAB.
We used the modules for LBP, made available with UGM\footnote{These
codes can be downloaded from \href{http://www.di.ens.fr/\textasciitilde mschmidt/Software/UGM.html}{UGM.html}
\href{http://www.di.ens.fr/\textasciitilde mschmidt/Software/UGM.html}{UGM.html}.} package. The LDA
models are learned using the lda package.\footnote{\url{http://chasen.org/\textasciitilde daiti-m/dist/lda/}.}

\textit{Threshold selection $r$ for our method.}
Recall that the parameter $r$ in our method controls the size of
neighborhoods over which the local MSTs are constructed in the first
step of our method. We earlier presented ranges of~$r$, where recovery
of the loopy structure is theoretically guaranteed (w.h.p.). However,
in\vadjust{\goodbreak}
practice, this range is unknown, since the model parameters are unknown
to the learner, and also since there is no ground truth with respect to
real datasets. Here, we present intuitive criterion for selecting the
threshold based on the BIC score. We choose the range for threshold $r$
as
%
\begin{equation}
\label{eqnrminmax} r_{\max} := \max_{(i,j) \in V\times V} d(i,j) ,\qquad
r_{\min} := \max_{j \in V} \min_{i \in V} d(i,j),
\end{equation}
thereby disallowing disconnected components in the output graph. Note
that if we choose $r \geq r_{\max}$, then the output is a latent tree.
In our experiments, we choose one value above $r_{\max}$ to find a
reference tree model and compare it with other outcomes. For the 20
newsgroup dataset, we find that $r_{\min} = 2.3678$ and $r_{\max} =
12.2692$. Therefore, we choose $r\in\{3,5,7,9,11,13\}$ for our
experiments on newsgroup data. For the monthly stock returns data,
$r_{\min} = 1.0337$ and $r_{\max} = 8.1172$, and we choose $r$ from
$1.1$ to $8.2$.
The tuning of parameters $\Lambda$ and $\tau$ has been previously
discussed in the context of learning latent trees (e.g.,~\cite{Choi&etal10JMLR}, page~1796), and we leverage on those results here.

\textit{Performance evaluation.} We evaluate performance based on the
test perplexity~\cite{newmanimproving} given by
%
\begin{equation}
\perpll:=\exp \Biggl[-\frac{1}{np}\sum_{k=1}^n
\log P\bigl({\mathbf x}^{\test}(k)\bigr) \Biggr],\label {eqnperpll}
\end{equation}
where $n$ is the number of test samples and $p$ is
the number of observed variables (i.e., words). Thus the perplexity is
monotonically decreasing in the test likelihood and a lower perplexity
indicates a better generalization performance. Along the lines of
\eqref
{eqnperpll}, we also evaluate the predictive perplexity~\cite{blei2003latent}
%
\begin{equation}
\predperpll:=\exp \Biggl[-\frac{1}{np}\sum_{k=1}^n
\log P\bigl({\mathbf x}^{\test}_{\pred}(k) \vert{\mathbf
x}^{\test
}_{\obs}(k)\bigr) , \Biggr]\label{eqnpredperpll}
\end{equation}
where a subset of word occurrences ${\mathbf x}^{\test}_{\obs}$ is
observed in
test data, and the performance of predicting the rest of words is
evaluated. In our experiments, we randomly select half the words in
test samples.

We also consider regularized versions of perplexity that capture
trade-off between model complexity and likelihood, given by
%
\begin{equation}
\perpbic:=\exp \biggl[-\frac{1}{np}\BIC\bigl({\mathbf x}^{\test}
\bigr) \biggr],\label {eqnperpbic}
\end{equation}
where the BIC score~\cite{schwarz1978estimating} is
defined as
%
\begin{equation}
\BIC\bigl({\mathbf x}^{\test}\bigr):= \sum_{k=1}^n
\log P\bigl({\mathbf x}^{\test}(k)\bigr)- 0.5 (\df)\log n ,
\end{equation}
where $\df
$ is the degrees of freedom in the model. For a graphical model, we set
$\df^{\GM}:= m+ |E|$, where $m$ is the total number of variables (both
observed and hidden), and $|E|$ is the number of edges in the model.\vadjust{\goodbreak}
For the LDA model, we set $\df^{\LDA}:= (p(m-p)-1)$, where $p$ is the
number of observed variables (i.e., words) and $m-p$ is the number of
hidden variables (i.e., topics). This is because a LDA model is
parameterized by a $p\times(m-p)$ topic probability matrix and a
$(m-p)$-length Dirichlet prior. Thus, the BIC perplexity in \eqref
{eqnperpbic} is monotonically decreasing in the BIC score, and a lower
BIC perplexity indicates better trade-off between model complexity and
data fitting. However, the likelihood and BIC score in \eqref
{eqnperpll} and \eqref{eqnperpbic} are not tractable for exact
evaluation in general graphical models since they involve the partition
function. We employ loopy belief propagation (LBP) to evaluate
them.\footnote{The likelihood is evaluated using $P({\mathbf
x}_V)=\frac{P({\mathbf x}_{V\cup H})}{P({\mathbf x}_H \vert{\mathbf x}_V)}$, where $P({\mathbf
x}_H \vert{\mathbf x}_V)$
and $P({\mathbf x}_{V\cup H})$ are computed using LBP, which is exact for
trees. The above expression holds for any configuration of hidden
variables ${\mathbf x}_H$, however we use the most likely hidden state to
avoid numerical issues.} Note that it is exact on a tree model and
approximate for loopy models. Along the lines of predictive perplexity
in \eqref{eqnpredperpll}, we also consider its regularized
version
%
\begin{equation}
\predperpbic:=\exp \biggl[-\frac{1}{np}\BIC\bigl({\mathbf
x}^{\test
}_{\pred}\vert{\mathbf x}^{\test}_{\obs}
\bigr) \biggr]\label {eqnpredperpbic} ,
\end{equation}
where the conditional BIC score is given by
%
\begin{equation}
\BIC\bigl({\mathbf x}^{\test
}_{\pred}\vert{\mathbf
x}^{\test}_{\obs}\bigr):= \sum_{k=1}^n
\log P\bigl({\mathbf x}^{\test}_{\pred}(k) \vert{\mathbf
x}^{\test}_{\obs
}(k)\bigr)- 0.5 (\df)\log n.
\end{equation}

In addition, we also evaluate topic coherence, frequently considered in
topic modeling. It is based on the average pointwise mutual information
(PMI) score
%
\begin{eqnarray}\label{eqnPMI}
\overline{\PMI}&:=&\frac{1}{45|H|}\sum_{h\in H}
\mathop{\sum_{i,j \in\Ac(h)}}_{i<j}
\PMI(X_i;X_j),
\nonumber
\\[-8pt]
\\[-8pt]
\nonumber
 \PMI (X_i;X_j)
&:=& \log\frac{P(X_i=1, X_j=1)}{P(X_i=1) P(X_j=1)},
\end{eqnarray}
where the set $\Ac(h)$ represents the ``top-10'' words associated with
topic $h\in H$. The number of such word pairs for each topic is
$\bigl({10 \atop2}\bigr)=45$, and is used for normalization. In \cite
{Newman&Karimi&Cavedon09ADCS}, it is found that the PMI scores are a
good measure of human evaluated topic coherence when it is computed
using an external corpus. It is also observed that using a related
external corpus gives a high PMI. Hence, in our experiments, we choose
a corpus containing news articles from the NYT articles bag-of-words
dataset. This dataset has a vocabulary of 102,660 words from 300,000
separate articles~\cite{Frank+Asuncion2010}. For LDA models, the top
10 words for each topic are selected based on the topic probability
vector. For latent graphical models, we use the criterion of
information distances on the learned model to select the 10 nearest
words for each topic.

\begin{figure}

\includegraphics{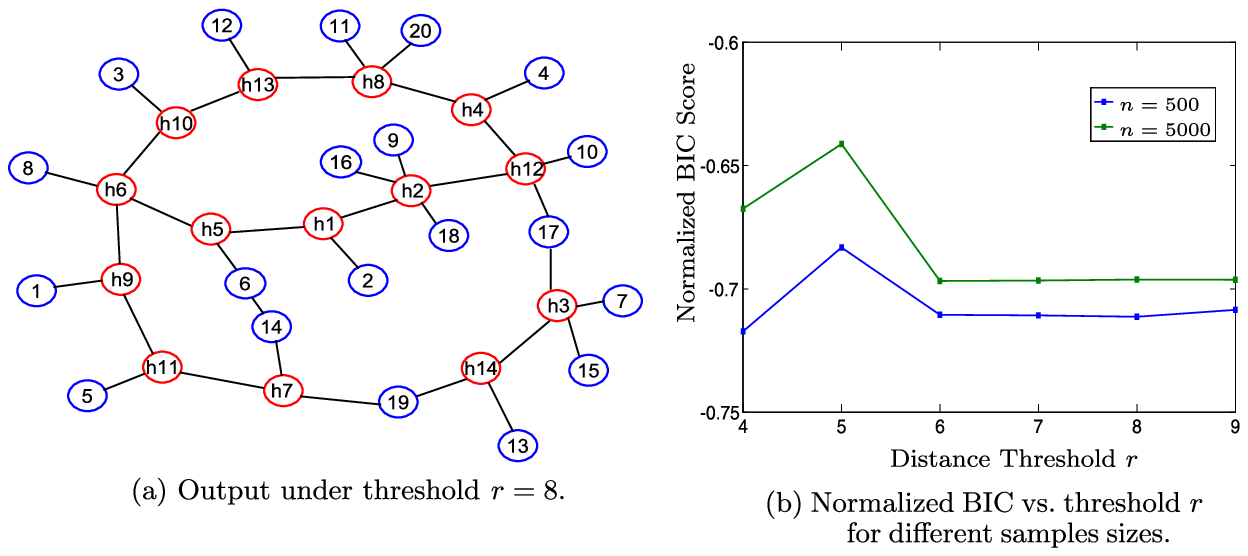}

\caption{Results for synthetic data with girth $g=10$ using the proposed method.}\label{figBICThresh}
\end{figure}


\subsection{Experimental results}
\mbox{}

\textit{Results for synthetic data}. We observe that
our method outputs graphs with a similar number of latent variables as
the ground truth when $r$ is chosen close to the bound $r_{\max}$,
defined in \eqref{eqnrminmax}. On the other hand, lower values of $r$
lead to more cycles and hidden variables in the output graph. The
normalized BIC scores (normalized with respect to $n$ and $p$) of the
loopy graphs improve with the number of samples $n$, as shown in
Figure~\ref{figBICThresh}(b). This is expected since the data becomes
less noisy with more samples. Figure~\ref{figBICThresh}(b) shows an
overall improvement in the normalized BIC score with increasing number
of samples $n$ for different thresholds~$r$. Figure
\ref{figBICThresh}(b)
shows the variation of normalized BIC scores for graphs learned using
thresholds $r=4$ to $9$ with girth $g=10$. We observe that the
normalized BIC score decreases for the lowest threshold $(r=4)$, where
the output graph shows a significant increase in latent nodes and
edges, resulting in overfitting, and higher thresholds have better BIC.
However, once the threshold results in a tree model, the BIC degrades
since the cycles are no longer present.

\begin{sidewaysfigure}

\centering\includegraphics{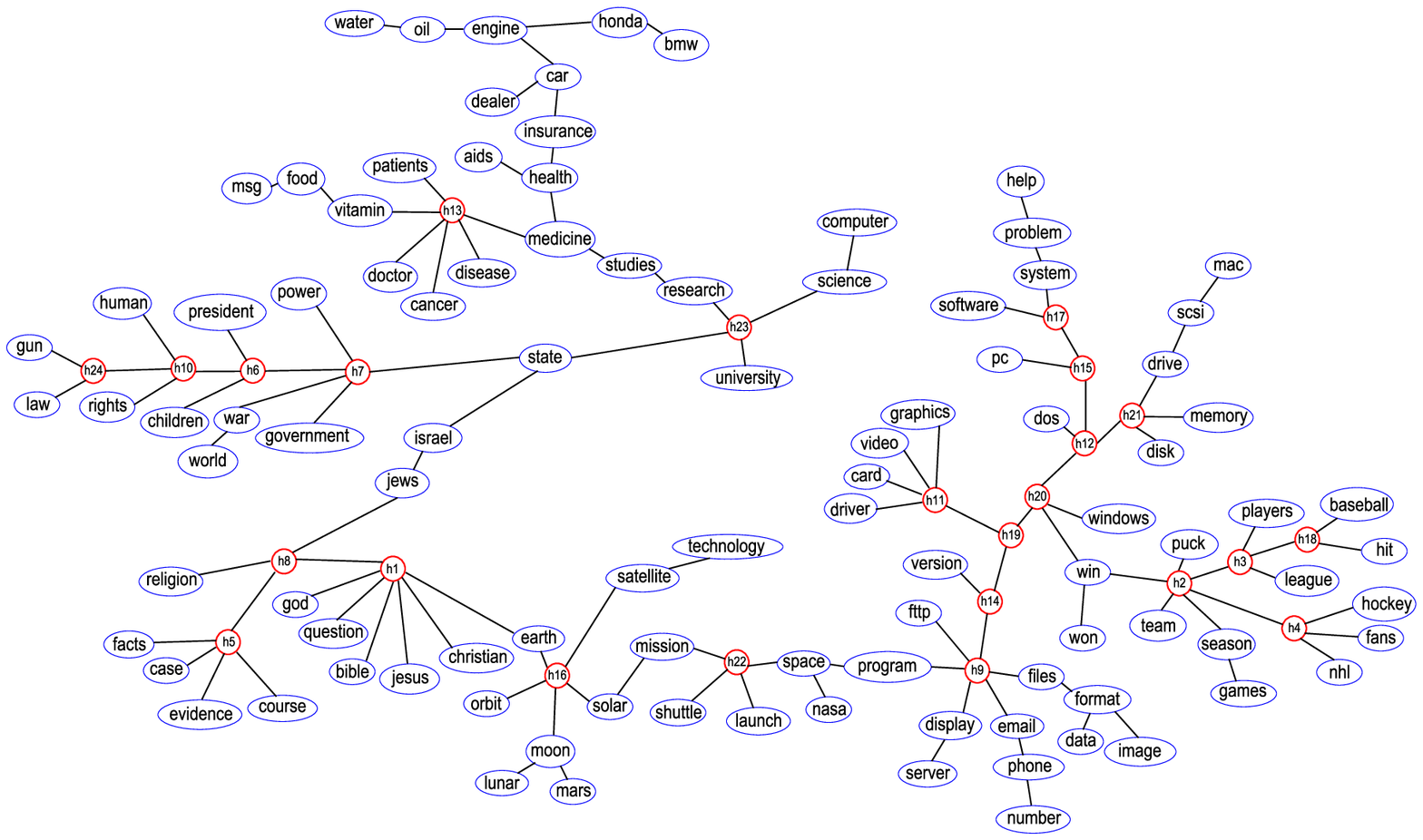}

\caption{Tree Graph Learned using $r=13$ with $\esttwo$ on 20 newsgroup
data.} \label{figdensityNewsgroup}
\end{sidewaysfigure}

\begin{sidewaysfigure}

\centering\includegraphics{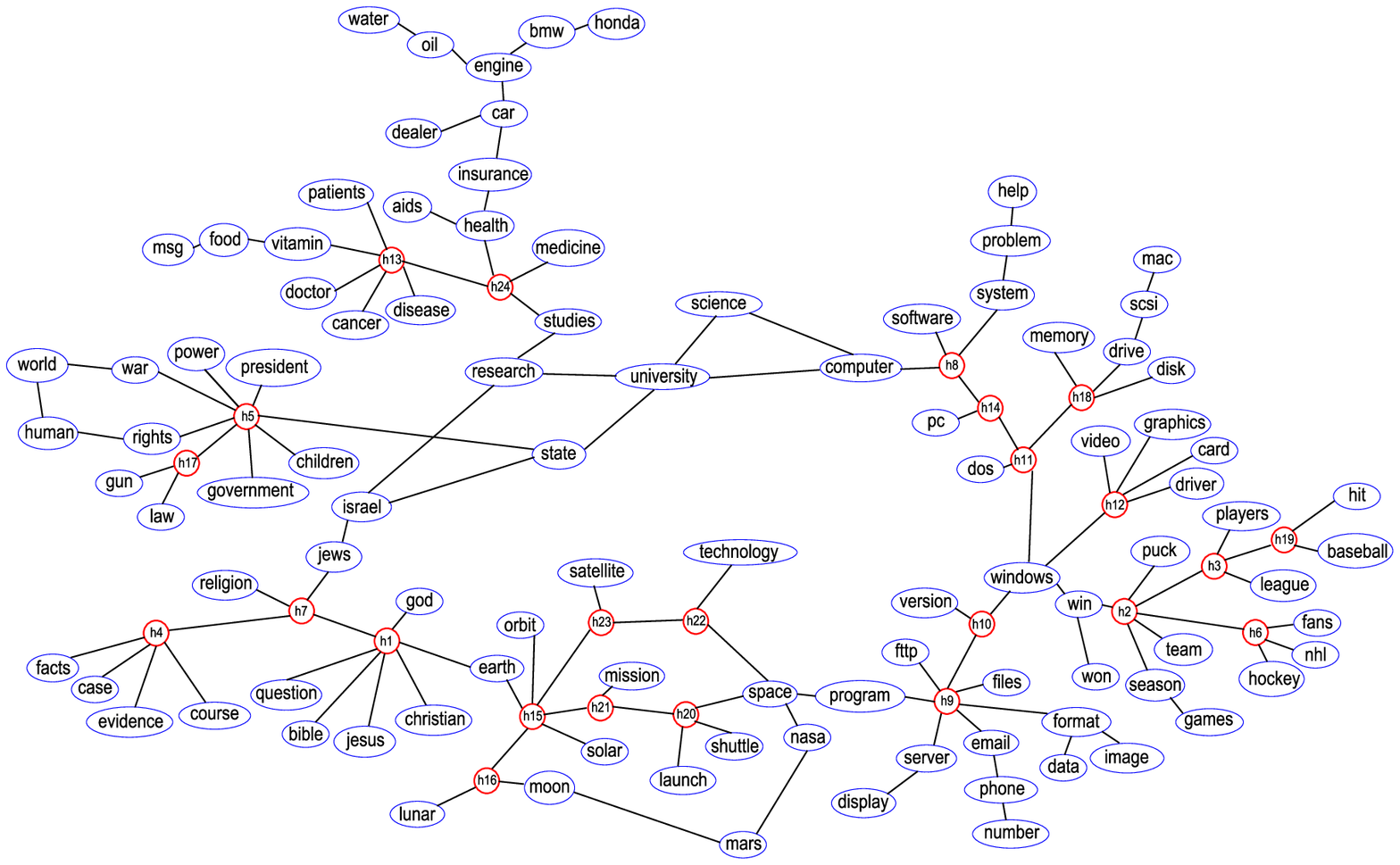}

\caption{Loopy Graph Learned using $r=9$ with $\esttwo$ on 20 newsgroup
data.}\label{figloopyR9New}
\end{sidewaysfigure}

\begin{table}
\tabcolsep=0pt
\caption{Comparison of proposed method under different thresholds $(r)$
with LDA under different number of topics (i.e., number of hidden
variables) on 20 newsgroup data. For definition of perplexity and
predictive perplexity based on test likelihood and BIC scores, and PMI,
see \protect\eqref{eqnperpll}, \protect\eqref{eqnpredperpll},
\protect\eqref{eqnperpbic},
\protect\eqref{eqnpredperpbic} and \protect\eqref{eqnPMI}}
\label{tabregLCLRGResultNew}
\begin{tabular*}{\textwidth}{@{\extracolsep{\fill}}lcccccccc@{}}
\hline
\multicolumn{1}{@{}l}{\textbf{Method}}&
\multicolumn{1}{c}{$\bolds{r}$} & \textbf{Hidden} & \textbf{Edges} & \textbf{PMI} & \textbf{Perp-LL} & \textbf{Perp-BIC} & \textbf{Pred-Perp-LL}
&\multicolumn{1}{c@{}}{\textbf{Pred-Perp-BIC}} \\
\hline
Proposed&
\phantom{0}3 & 55 & 265 & 0.2638 & 1.1533 & 1.1560 & 1.0695 & 1.0720 \\
Proposed&
\phantom{0}5 & 39 & 293 & 0.4875 & 1.1567 & 1.1594 & 1.0424 & 1.0448 \\
Proposed&
\phantom{0}7 & 32 & 183 & 0.4313 & 1.1498& 1.1518 & 1.0664 & 1.0682\\
Proposed&
\phantom{0}9& 24 & 129 & 0.6037 & 1.1543 & 1.1560 & 1.0780 & 1.0795\\
Proposed&
11& 26 & 125 & 0.4585 & 1.1555 & 1.1571& 1.0787 & 1.0802\\
Proposed&
13& 24 & 123& 0.4289 & 1.1560 & 1.1576 & 1.0788 & 1.0803\\
LDA&NA& 10 & NA & 0.2921 & 1.1480 & 1.1544 &1.1623 & 1.1656 \\
LDA&NA& 20 & NA & 0.1919 & 1.1348 & 1.1474 & 1.1572 & 1.1638 \\
LDA&NA& 30 & NA & 0.1653 & 1.1421 & 1.1612 & 1.1616 & 1.1715\\
LDA&NA& 40 & NA & 0.1470 & 1.1494 & 1.1752& 1.1634 & 1.1767\\
\hline
\end{tabular*}
\end{table}

\textit{Graph structure for newsgroup data}. We employ our method to
learn the graph structures under different thresholds $r\in\{
3,5,7,9,11,13\}$ on newsgroup data, which controls the length of
cycles. At $r=13$ as shown in Figure~\ref{figdensityNewsgroup}, we obtain a
latent tree, and for $r\in\{3,5,7,9\}$, we obtain loopy models. The
first long cycle appears at $r=9$ shown in Figure~\ref{figloopyR9New}.
At $r=7$, we find a combination of short and long cycles. We find that
models with cycles are more effective in discovering intuitive
relationships. For instance, in the latent tree $(r=13)$, the link
between ``computer'' and ``software'' is missing due to the tree
constraint, but is discovered when $r\leq9$. Moreover, we see that
common words across different topics tend to connect the local
subgraphs. For instance, the word ``program'' is used in the context of
both space program and computer programs. Similarly, the word ``earth''
is used both in the context of religion and space exploration.

\textit{Perplexity and topic coherence for newsgroup data}. In
Table~\ref{tabregLCLRGResultNew}, we present results under our method
and under LDA modeling on newsgroup data. For the LDA model, we vary
the number of hidden variables (i.e., topics) as $\{10, 20, 30, 40\}$.
In contrast, our method is designed to optimize for the number of
hidden variables, and does not need this input. We note that our method
is competitive in terms of both predictive perplexity and topic
coherence. We find that the topic coherence (i.e., PMI) for our method
is optimal at $r=9$, where the graph has a single long cycle and a few
short cycles. Intuitively, this model is able to discover more
relationships between words, which the latent tree $(r=13)$ is unable
to do so. On the other hand, for $r < 9$, topic coherence is degraded
which suggests that adding too many cycles is counterproductive.
However, the model at $r=5$ performs better in terms of predictive
perplexity indicating that it is able to use evidence from more
observed words for prediction on test data. Moreover, all of our latent
graphical models outperform the LDA models in terms of predictive
perplexity. The top 10 topic words for selected topics are given for
our method at $(r=9)$ and for the LDA model (with 10 topics) are given
in Tables~\ref{tabtopicsR9} and~\ref{tabtopicsLDAK10}.

\begin{table}
\caption{Top 10 topic words from selected topics in loopy graphical
model\break with threshold $r=9$, the topic number corresponds to the
labels\break
of hidden variables in the loopy graph shown in Figure \protect\ref
{figloopyR9New}} \label{tabtopicsR9}
\begin{tabular*}{\textwidth}{@{\extracolsep{\fill}}lcccc@{}}
\hline
\textbf{Topic 16} & \textbf{Topic 18} & \textbf{Topic 12} & \textbf{Topic 1} & \textbf{Topic 8} \\
\hline
lunar & disk & card & god & software \\
moon & drive & video & jesus & pc \\
orbit & dos & windows & bible & computer\\
solar & memory & driver & christian & system\\
mission & windows & graphics & religion & dos\\
satellite & pc & dos & earth & windows \\
earth & software & version & question & disk\\
shuttle & scsi & ftp & fact & science\\
mars & computer & pc & jews & drive \\
space & system & disk & evidence & university\\
\hline
\end{tabular*}
\end{table}

\begin{table}[b]
\caption{Top 10 topic words corresponding to selected topics from the
LDA model with 10 topics} \label{tabtopicsLDAK10}
\begin{tabular*}{\textwidth}{@{\extracolsep{\fill}}lcccc@{}}
\hline
\textbf{Topic 4} & \textbf{Topic 8}& \textbf{Topic 7} & \textbf{Topic 6} & \textbf{Topic 5} \\
\hline
Space & windows & card & god & drive \\
nasa & files & graphics& world & states \\
insurance & dos & video & fact & research \\
earth & format & driver & christian & disk \\
moon & ftp& windows & jesus & university \\
orbit & program & computer & religion & mac \\
mission & software & pc & bible & scsi \\
launch & win & version & evidence & computer \\
gun & version & software & human & system \\
shuttle & pc & system & question & power\\
\hline
\end{tabular*}
\end{table}

\begin{table}
\caption{Comparison of proposed method under different thresholds $(r)$
on Stock data\break using the proposed method. For definition of perplexity
based on test likelihood\break and BIC scores; see \protect\eqref
{eqnperpll} and
\protect\eqref{eqnperpbic}} \label{tabLCLRGStockResultNew}
\begin{tabular*}{\textwidth}{@{\extracolsep{\fill}}lcccc@{}}
\hline
\multicolumn{1}{@{}l}{$\bolds{r}$} & \textbf{Hidden} & \textbf{Edges} & \textbf{Perp-LL} & \textbf{Perp-BIC} \\
\hline
2.7 & 35 & 154 & 1.9498& 2.0296 \\
3.9 & 39 & 139 & 2.0200 & 2.0993\\
4.9 & 35 & 129 & 2.0210 & 2.0960\\
5\phantom{0.} & 36 & 131& 2.0169 & 2.0927\\
6.7 & 26 & 111& 2.0344 & 2.1016\\
7.7 & 26 & 111& 2.0353 & 2.1025\\
8.2 & 26 & 110& 2.0405 & 2.1076\\
\hline
\end{tabular*}
\end{table}

\begin{sidewaysfigure}

\centering\includegraphics{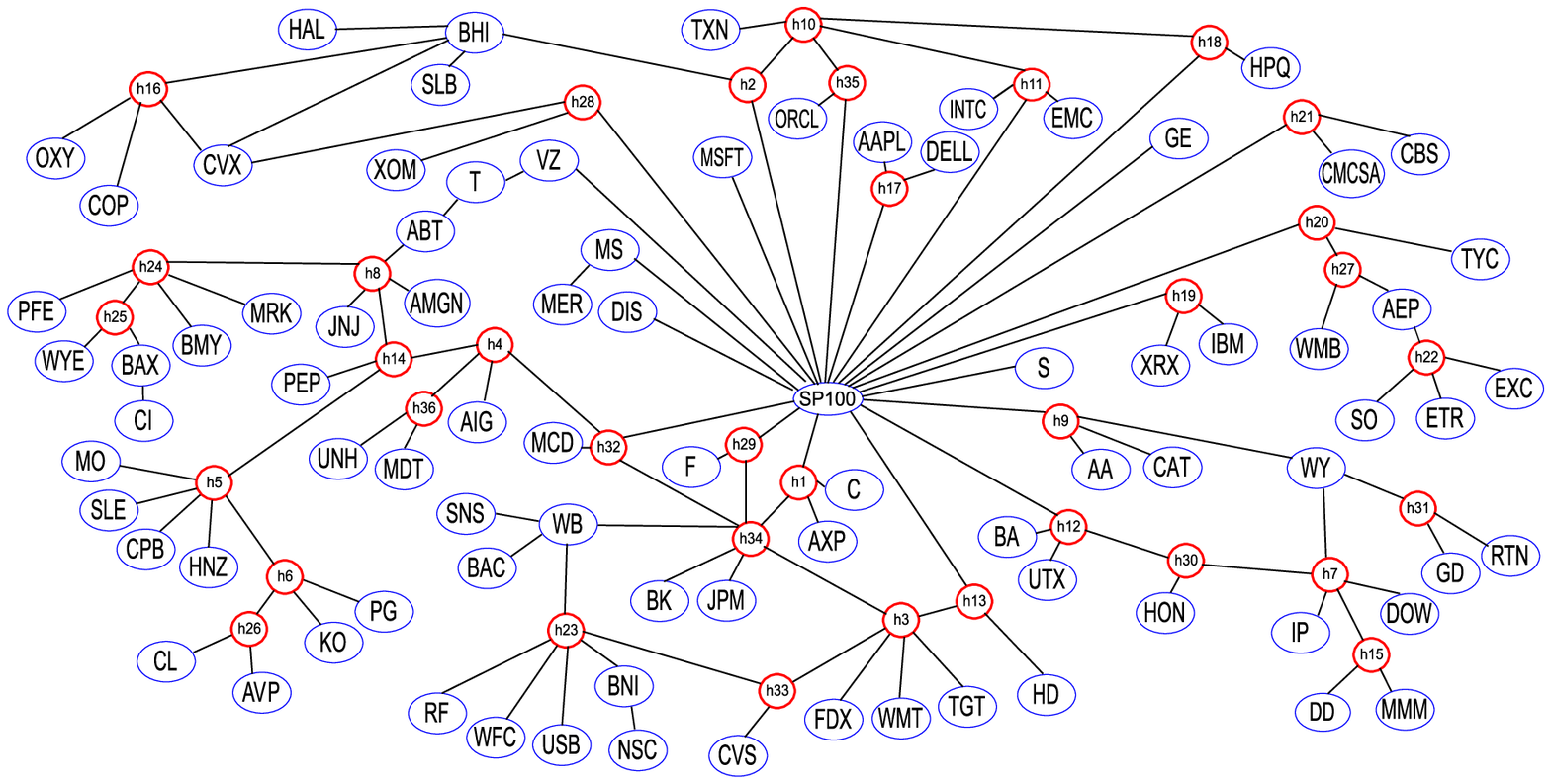}

\caption{Loopy Graph Learned using $r=5$ with $\estone$ on S$\&$P 100
monthly stock return data.}\label{figTreeNew}
\end{sidewaysfigure}

\begin{sidewaysfigure}

\centering\includegraphics{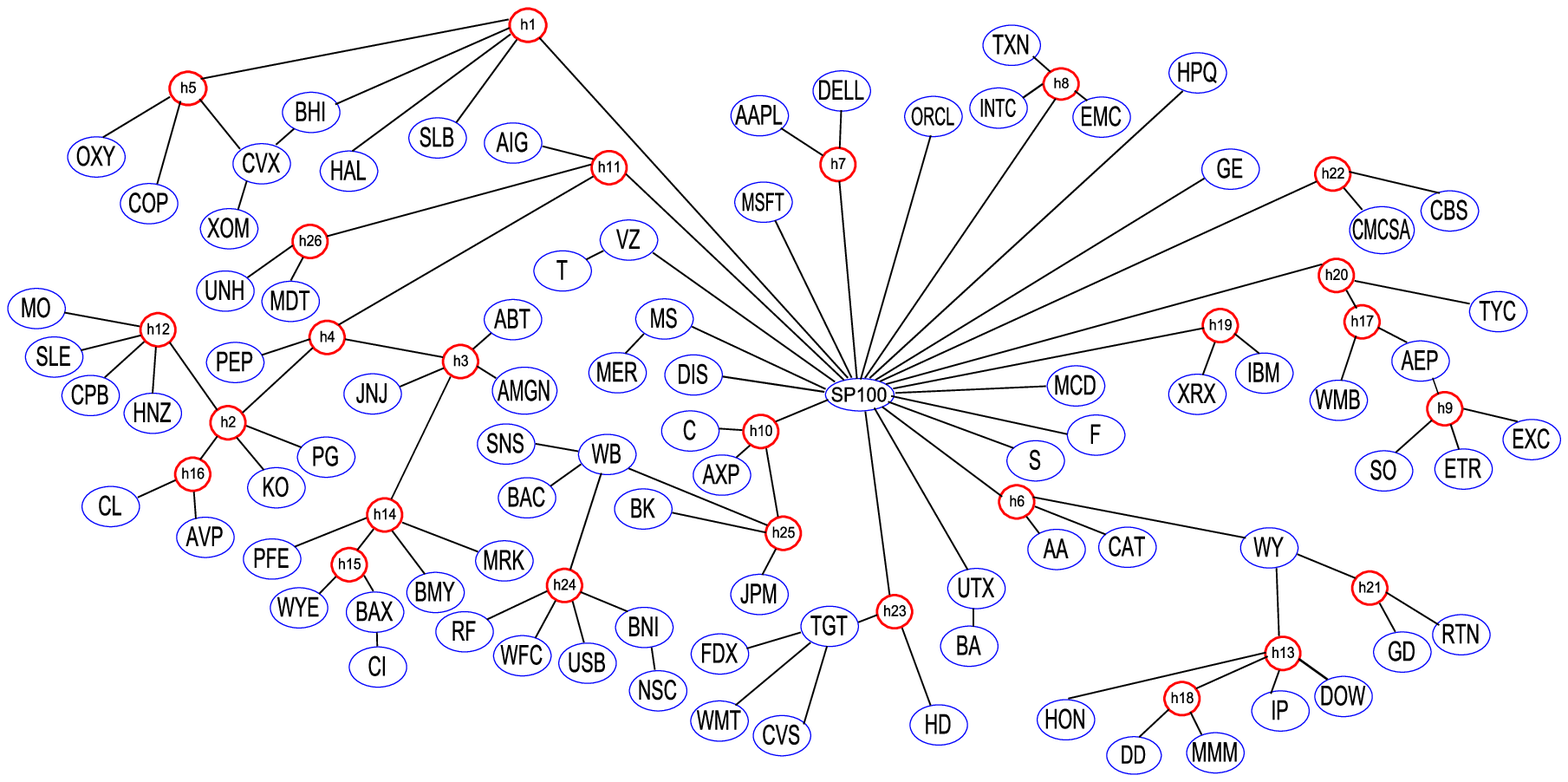}

\caption{Loopy Graph Learned using $r=7.7$ with $\estone$ on S$\&$P 100
monthly stock return data.}\label{figStockloop5}
\end{sidewaysfigure}

\begin{sidewaysfigure}

\centering\includegraphics{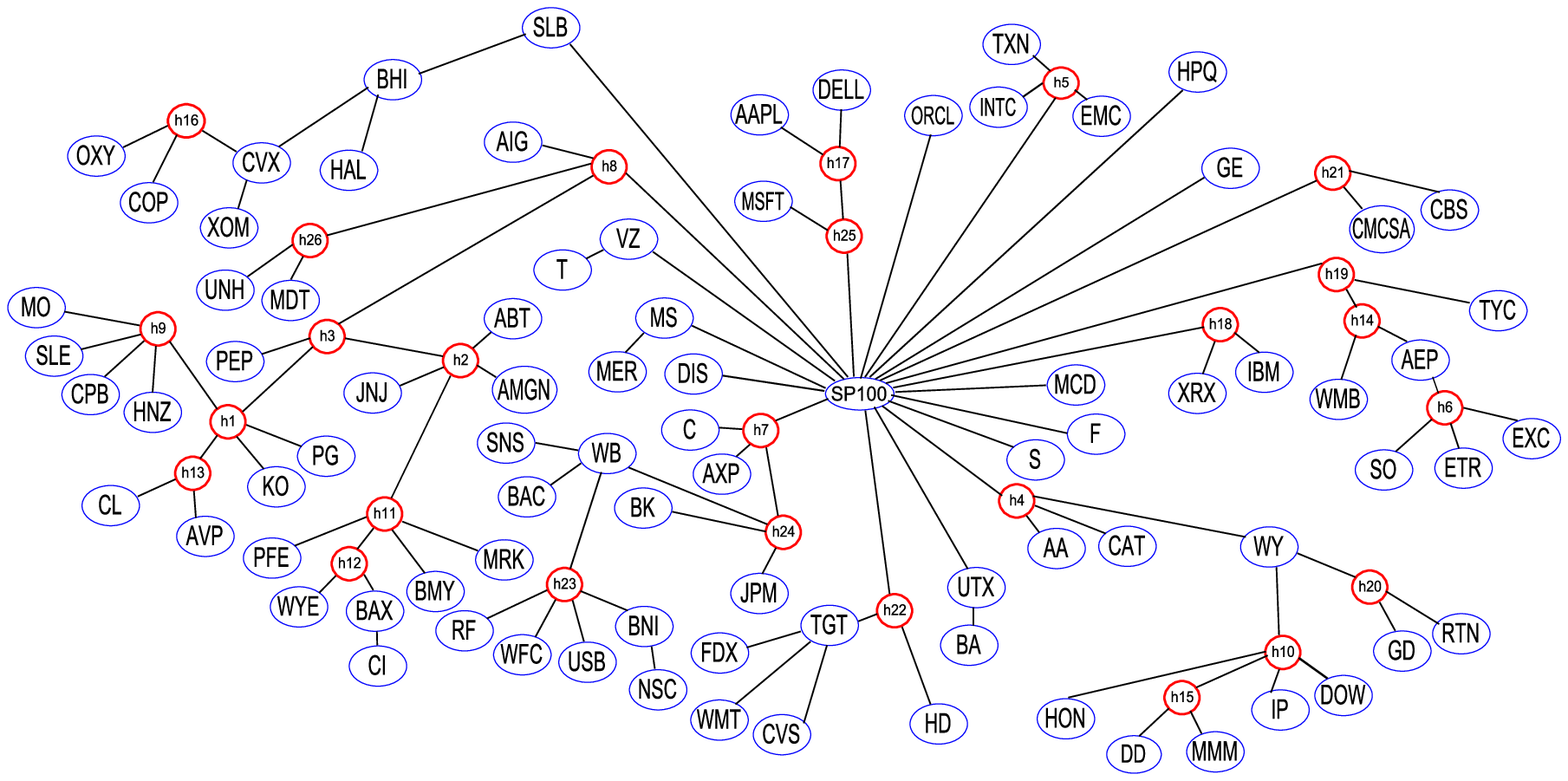}

\caption{Tree Graph Learned using $r=8.2$ with $\estone$ on S$\&$P 100
monthly stock return data.}
\label{figStockloop77}
\end{sidewaysfigure}

\textit{Graph structure for stock market data}.
The outcome of applying the proposed algorithm to stock market data is
presented in Table~\ref{tabLCLRGStockResultNew}. We observe that the
number of edges and hidden variables remain fairly constant over a
large range of thresholds. Specifically for $r \in[5.9, 6.7] \cup
[6.8,7.7]$, we obtain the same graph structure (for $r > r_{\max}$, we
obtain a tree).
Another general trend observed is the improvement of
the BIC score as the threshold decreases up till a certain point. The
graphs learned using $r= 5,  7.7 $ and $8.2$ are shown in Figures \ref
{figTreeNew},~\ref{figStockloop5} and~\ref{figStockloop77}.
Interesting connections between companies emerge. The latent tree
structure in Figure~\ref{figStockloop77} captures many key
relationships. In particular, the S\&P index node has a high degree
since it captures the overall trend of various companies. Companies in
similar sectors and divisions are grouped together. For instance,
retail stores such as ``Target,'' ``Walmart,'' ``CVS'' and ``Home
Depot'' are grouped together. However, additional relationships emerge
as the threshold is decreased and cycles are added. We observe that the
first cycle that is added connects the various oil companies which
suggests strong interdependencies and influence on the S\&P 100 index.
In addition, more cycles emerge when the threshold is decreased
further.
For instance, in Figure~\ref{figTreeNew}, we find a cycle
connecting aviation company ``Boeing'' with ``Honeywell'' which is in
the aviation industry, but also additionally is in the chemical
industry and connects to companies such as ``Dow Chemicals.'' Thus as
in newsgroup data, we find that companies in multiple categories lead
to cycles in the underlying graph.

%
\begin{figure}

\includegraphics{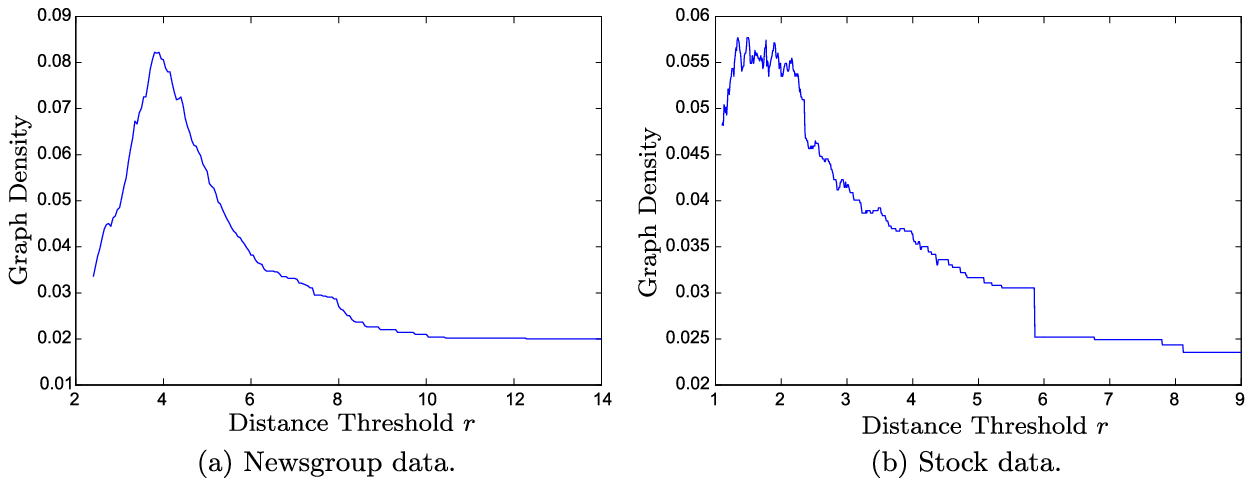}

\caption{Variation of edge density of graphs at the initialization
stage of $\estone$ vs. threshold $r$.}
\label{figtreeStock}
\end{figure}

\textit{Edge density vs. threshold $r$}. We now study the edge
density (i.e., number of edges) in the initialization step of our method
as a function of the threshold $r$ for both newsgroup and stock data.
Recall that the initialization step involves building a loopy graph on
observed variables (and no hidden variables). The edge density in this
step is indicative of the number of cycles added to the ultimate latent
model. We observe that the graphs become denser as $r$ is reduced from
$r_{\max}$. However, when $r$ is very small, the number of edges
decreases since the nodes have sparser neighborhoods. This trend is
seen in both Figures~\ref{figtreeStock}(a) and~\ref{figtreeStock}(b) which show the variation for newsgroup and stock
data. For the newsgroup data, the graph density peaks at $r=5$, which
also achieves the highest predictive perplexity; see Table \ref
{tabregLCLRGResultNew}. Thus, we see a direct relationship between the
edge density and the corresponding predictive perplexity in the learned
model. Intuitively, this is because as the number of edges increases,
prediction at any node involves more evidence. However, as the
threshold $r$ is reduced further, graphs become less denser, and there
is also a corresponding degradation in the predictive perplexity.



%

The above experiments confirm the effectiveness of our approach for
discovering hidden topics and are in line with the theoretical
guarantees established earlier in the paper. Our analysis reveals that
a large class of loopy graphical models with latent variables can be
learned efficiently in different domains.

\section{Conclusion}
In this paper, we considered latent graphical models Mar\-kov on
girth-constrained graphs and proposed a novel approach for structure
estimation. We established the correctness of the method when the model
is in the regime of correlation decay and also derived PAC learning
guarantees. We compared these guarantees with other methods for
graphical model selection, where there are no latent variables. Our
findings reveal that latent variables do not add much complexity to the
learning process in certain models and regimes, even when the number of
hidden variables is large. These findings push the realm of tractable
latent models for learning.

\section*{Acknowledgments}
The authors thank E. Mossel (Berkeley) for detailed discussions in the
beginning regarding problem formulation, modeling and algorithmic
approaches and Padhraic Smyth (UCI) and David Newman (UCI) for
evaluation measures for topic models. The authors also thank the editor
Tony Cai (Wharton) and anonymous reviewers whose comments substantially
improved the paper. An abridged version of this work appears in the
Proceedings of NIPS 2012.

\begin{supplement}[id=suppA]
\stitle{Supplementary material to ``Learning loopy graphical models with latent variables:
Efficient methods and guarantees''}
\slink[doi,text={10.1214/12-\break AOS1070SUPP}]{10.1214/12-AOS1070SUPP} 
\sdatatype{.pdf}
\sfilename{aos1070\_supp.pdf}
\sdescription{Proofs of various theorems.}
\end{supplement}


%

%

%

%

%

\printaddresses

\end{document}